\documentclass{article}

\usepackage{arxiv}

\usepackage[utf8]{inputenc}
\usepackage[T1]{fontenc}
\usepackage{url}
\usepackage{booktabs}
\usepackage{amsfonts}
\usepackage{amsmath}
\usepackage{nicefrac}
\usepackage{microtype}
\usepackage{lipsum}
\usepackage{graphicx}
\usepackage{natbib}
\usepackage{doi}
\usepackage{tikz}
\usepackage{subfig}
\usepackage{titlesec}
\usepackage{ccicons}
\usepackage{longtable}
\usepackage{multicol}
\usepackage{multirow}
\usepackage{rotating}
\usepackage{makecell}
\usepackage[threshold=8,autopunct,csdisplay,autostyle]{csquotes}
\usepackage{array,ragged2e}
\usepackage{outlines}
\usepackage{tabularx}
\usepackage{mdframed}
\usepackage{enumitem}
\usepackage{rotating}
\usepackage{accessibility}
\usepackage{hyperref}
\hypersetup{hidelinks}
\usepackage{cleveref}

\providecommand{\Description}[2][]{}

\newcommand{\todo}[1]{} 
\newcommand{\query}[1]{} 
\newcommand{\queryyy}[1]{}
\newcommand{\toinclude}[1]{}

\title{Designing the Future of User Feedback for Generative AI}

\date{}

\newif\ifuniqueAffiliation
\uniqueAffiliationtrue

\usepackage{authblk}

\newbox{\orcid}
\sbox{\orcid}{\includegraphics[scale=0.06]{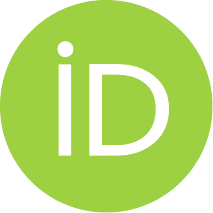}}

\author[1,2]{%
  \href{https://orcid.org/0000-0002-8477-7276}{%
    \usebox{\orcid}\hspace{1mm}Alisa Frik%
  }\thanks{\texttt{afrik@icsi.berkeley.edu}}%
}

\author[1,2]{%
  \href{https://orcid.org/0000-0002-9792-2743}{%
    \usebox{\orcid}\hspace{1mm}Julia Bernd%
  }\thanks{\texttt{jbernd@icsi.berkeley.edu}}%
}

\author[3]{%
  Amitis Karami\thanks{\texttt{akarami@ebay.com}}%
}

\author[1,3]{%
  \href{https://orcid.org/0000-0001-9666-2663}{%
    \usebox{\orcid}\hspace{1mm}Mohammad Tahaei%
  }\thanks{\texttt{mtahaei@ebay.com}}%
}

\affil[1]{International Computer Science Institute, Berkeley, CA, USA}
\affil[2]{University of California, Berkeley, Berkeley, CA, USA}
\affil[3]{eBay Inc., San Jose, CA, USA}

\renewcommand{\headeright}{}
\renewcommand{\undertitle}{}
\renewcommand{\shorttitle}{Designing the Future of User Feedback for Generative AI}

\begin{document}
\maketitle

\begin{abstract}
Post-deployment feedback from users can be a cost-effective, scalable, and representative means to monitor and improve 
generative AI systems and features. 
When implemented effectively, giving such feedback can increase users' engagement with and trust in GenAI systems. 
Government regulations and industry guidelines call for post-deployment user engagement, but there is little guidance on designing mechanisms that 
are \textit{usable} for consumers and provide \textit{actionable} input for product teams.

We conducted a multi-phase study as a collaboration between academic researchers and eBay.
Our benchmark evaluation of current industry approaches 
identified common issues 
including lack of discoverability, 
unclear terminology, and inattention to user value. Based on these findings, we developed best-practice recommendations 
and designed and tested a prototype feedback-collection 
tool. The tool aimed to provide users with an efficient, flexible, 
and positive feedback-giving experience, and provide product teams with rich data on performance and 
potential 
problems in a usable format. 
\todo{Now 147 words according to Word. -JB // We need to make sure it stays under 150 words at the time of submission. - AF} 
\end{abstract}

\keywords{Generative AI \and Responsible AI \and User Feedback \and Interface Design \and Heuristic Evaluation}

\section{Introduction}
\label{sec: intro}

\begin{figure}[htb]
\begin{center}
\includegraphics[width=\columnwidth]{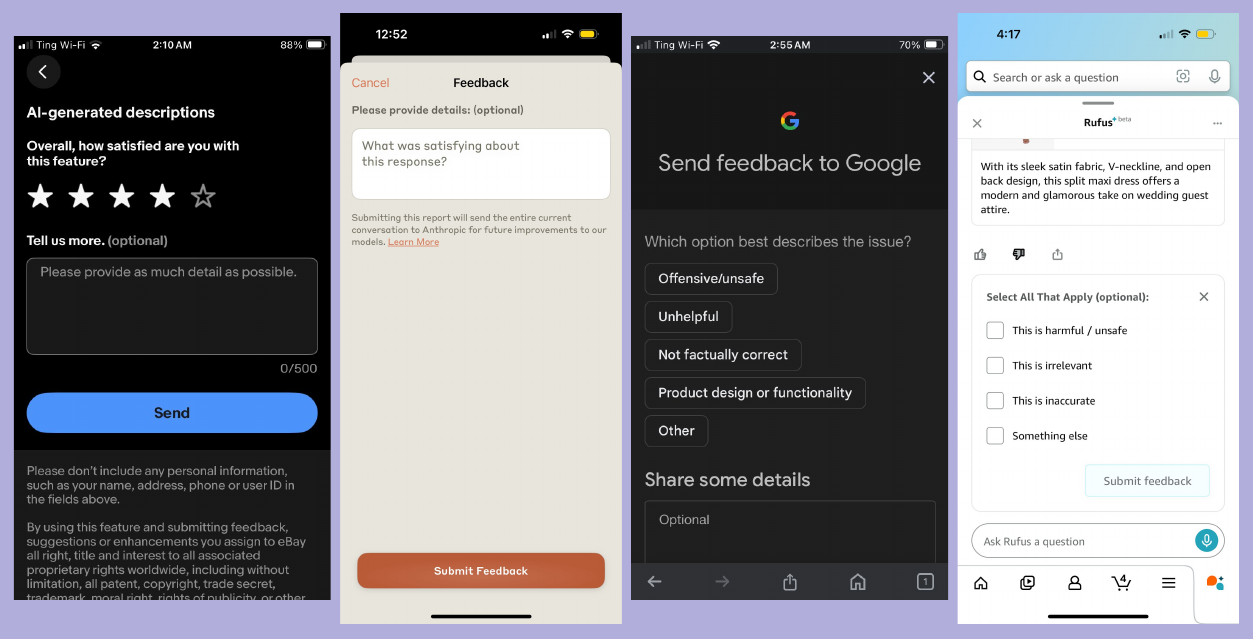} 
\end{center}
\caption{Example feedback forms for eBay's AI Description Generator, Claude, Google Search's AI Overview, and Amazon's Rufus. Screenshots are from March 2025.}
\Description[Forms showing types of ratings, headers used, and options to choose from for each.]{The form for eBay's AI Description Generator is headed ``AI-generated descriptions''; asks ``Overall, how satisfied are you with this feature?''; provides the option to choose up to five stars; and offers a textbox headed ``Tell us more''. The form for Claude is headed ``Feedback''; requests ``Please provide details''; and offers a textbox with ghost text saying ``What was satisfying about this response?'' The form for Google Search's AI Overview is headed ``Send feedback to Google''; asks ``Which option best describes the issue?''; provides selectable options for ``Offensive/unsafe'', ``Unhelpful'', ``Not factually correct'', ``Product design or functionality'', and ``Other''; and offers a textbox headed ``Share some details''. The form for Amazon's Rufus is invoked by a thumbs down; asks ``Select all that apply''; and provides selectable options for ``This is harmful/unsafe'', ``This is irrelevant'', ``This is inaccurate'', and ``Something else''. }
\label{fig:example_forms}
\end{figure}

Generative AI (GenAI) systems, especially Large Language Models (LLMs), have rapidly moved from research settings into everyday digital products \cite{sun2024generative}. 
GenAI capabilities are now embedded in conversational assistants and other interfaces that support content creation and information work \cite{openai2022chatgpt,tahaei2023hcerai}. Human feedback is central to how these systems are built and improved \cite{hashmati2025eXplainable}: it enters through data creation and annotation, supervised fine-tuning and reinforcement learning, and post-deployment interaction and evaluation \cite{nist2023airmf,autio2024genaiprofile}. In deployed products, feedback is therefore not a peripheral interface feature; it is part of how providers monitor system behavior and iteratively refine model-enabled systems \cite{nist2023airmf,autio2024genaiprofile}. Figure \ref{fig:example_forms} shows some examples of feedback forms for GenAI products and features.

User feedback, specifically (as opposed to feedback from trained annotators) offers cost-effective, scaled, contextual, and potentially more representative signal for refining model outputs. Moreover, ability to provide intentional feedback may increase users' engagement and trust, explicitly involving them in system improvements and fine-tuning of models, and it can boost a company's reputation and support efforts in addressing safety and ethical challenges.

Recent events provide evidence of why these feedback loops matter. In 2025, OpenAI reported that a GPT-4o update had made ChatGPT overly sycophantic---excessively obedient and insincerely flattering---and identified an additional reward signal informed in part by thumbs-up and thumbs-down feedback as one of the contributing factors \cite{openai2025sycophancy}. This illustrates that post-deployment feedback can shape model behavior, not merely customer support or surface-level user experience. Moreover, user feedback helps companies fine-tune models in a cost-effective manner by supplementing limited internal testing capacity with crowdsourced model-improvement signals. Major providers explicitly position feedback as part of improvement and monitoring~\cite{openai2025datause,google2025geminifeedback,microsoft2025copilotfeedback}. 

The policy context makes the need for effective monitoring especially salient. NIST’s AI Risk Management Framework and its Generative AI Profile frame AI risk management across the AI life cycle \cite{nist2023airmf,autio2024genaiprofile,nist2026monitoring}. The EU AI Act also requires post-market monitoring systems for high-risk AI systems that actively and systematically collect, document, and analyze relevant data throughout the system's lifetime \cite{eu2024aiact}. User feedback mechanisms could 
help operationalize post-deployment monitoring policy requirements for both high-risk and lower-risk systems.

However, while existing regulations and standards articulate governance objectives, concrete guidance about how to collect and process user feedback about GenAI features is currently limited in both industry and academia. There are no official standards or empirically grounded design frameworks for user feedback mechanisms that explain how feedback tools should be made discoverable, legible, privacy-aware, and actionable for both users and organizations. Nor is there much understanding of what users' actual experience with or perceptions of those feedback mechanisms are, or what kinds of feedback they would like to provide to companies in the first place. The lack of such standards and guidance limits companies' use of user feedback for fine-tuning of the models, which in turn hinders efforts to improve 1) model performance, 2) adherence to responsible AI practices, policies, and regulations (e.g., the AI EU Act), and 3) user experience, trust, and satisfaction with the AI-powered products. This makes end-user feedback tools an important but understudied layer of GenAI governance. 
Thus, the overarching objective for this research is to identify best practices for designing and implementing usable and useful user feedback tools for digital products that use GenAI features. 
We aim to answer the following research questions:
\begin{itemize}
 \item \textbf{RQ1:} How do different products with GenAI features currently collect feedback from users about those features?
 \item \textbf{RQ2:} What do researchers, product team stakeholders, and users view as 
 best practices and usability issues for  
 existing and proposed user-feedback collection mechanisms?
\end{itemize}

To answer these questions, we conducted a multi-phase study that benchmarked current industry approaches, assessed them through usability-oriented inspection and heuristic evaluation, and iteratively developed and refined a prototype with input from industry stakeholders and users. 
This work was conducted in collaboration with eBay, a large e-commerce platform
that makes extensive use of GenAI and related user feedback mechanisms. This collaboration grounds our contributions in an industry context and helps ensure that the resulting design guidelines are applicable and usable in real-world settings. Across these phases, we identified recurring issues in discoverability, clarity, consistency, transparency, and perceived user value, and translated those findings into an empirically grounded set of design recommendations for user feedback tools in GenAI systems.

The contributions of this research include: 
\begin{itemize}
    \item 
    A taxonomy of the current landscape of existing approaches to collecting user feedback about GenAI features used by eBay and other companies;
    \item An assessment of existing approaches via 
    heuristic evaluation; 
    \item A user-tested prototype feedback tool;
    \item 
    Design guidelines and best practices for feedback tools for GenAI-powered features.
\end{itemize}

\section{Related Work}
\label{sec:related_work}


\subsection{
User Feedback Mechanisms for Generative AI}
\label{relwork_aifeedback}
With the advent of ChatGPT and the proliferation of Generative AI (GenAI), the need to carefully design how systems communicate with and learn from users has intensified \cite{weisz2023toward,liao2023transparency,zamfirescu2023why,casper2023open}. 
Within HCI, it is well established that interfaces fundamentally mediate how users understand and interact with digital systems~\cite{norman1986cognitive,norman2013design}. Thus, user feedback mechanisms play an important role in closing the loop between AI outputs, user experience, and model improvement, making AI systems smarter, safer, and more user-centered. \citet{kirk2023past} surveyed 95 papers to summarize past and present approaches for using human feedback in machine learning for training and fine-tuning LLMs (including explicit comparisons of outputs by human raters, fine-grained feedback, edits, implicit feedback, and feedback collected from social media and forums), and discussed suggestions for future improvements to those approaches. 

While extensive machine learning research focuses on how human feedback can be used to train and fine-tune foundation models (e.g., via Reinforcement Learning from Human Feedback (RLHF), where the model generates several outputs and human raters vote on the preferred options based on specified criteria)~\cite[e.g.,][]{lin2020review,griffith2013policy,retzlaff2024human,ziegler2019fine,ouyang2022training,bai2022training}, there remains a critical gap regarding effective \textit{design patterns} that can be used to capture this data. 
Moreover, relying on binary votes, as is typical for RLHF, limits understanding of what aspects of the output influenced a rater's preference~\cite{wu2023fine}---though this may also be an issue even with more fine-grained feedback options \cite{collins2025beyond}. 
\textit{How} that feedback mechanism is designed is critical for its success, as prior research demonstrates that interface design and choice architecture directly impact user choices, significantly altering user perceptions and behaviors~\cite{thaler2008nudge,thaler2013choice,munscher2016review,johnson2012beyond}. 
Therefore, the feedback interface is not a neutral conduit; it is a critical socio-technical instrument that shapes the very systems that rely upon it to function and improve.

Moreover, prior research demonstrates that it is beneficial to “keep humans in the loop” to ensure adequate levels of control over ethical violations in AI systems~\cite{cranor2008framework} as well as adherence to ethical AI guidelines~\cite{chen2023constructing}. Additionally, preserving agency is an important value for humans when interacting with computing systems~\cite{sundar2010personalization,huang2023calling,tene2012big}. Concerns about the “alignment problem” in AI (i.e. concerns that AI will take away control from humans and will pose existential threats to humanity) also illustrate how important agency and a sense of control are for human users of AI~\cite{yudkowsky2016ai,gabriel2020artificial,ghazavichallenge}. Thus, we hypothesize that well-designed feedback tools could support a sense of agency, control, and trust among users of AI-powered features and services.

\subsection{Existing AI Governance Guidelines about Feedback for GenAI Systems}
Scrutiny of AI systems and structured efforts to mitigate risks posed by AI are increasing, with a growing number of governance and regulatory frameworks for AI emerging around the world~\cite[see overview in][]{radu2021steering,ulnicane2022governance}. The most prominent examples include the Chinese government's AI regulations~\cite[see overview in][]{franks2024report} and the European Union's Artificial Intelligence Act~\cite{AIEU}, which call for greater transparency and impose evaluations of whether AI systems are trustworthy and responsibly developed. Frameworks for multilateral governance of AI have also emerged to address advanced AI systems, including the G7 Code of Conduct~\cite{EU2023code}, the Guiding Principles for Advanced AI System~\cite{EU2023principles}, and the Bletchley Declaration~\cite{Bletchley2025}. 

To address the growing need to manage responsibility for AI safety and trustworthiness, businesses, government, non-profits, and civil society organizations have started developing various guidelines, taxonomies, and frameworks for evaluating and addressing ethical risks posed by AI. Notable examples include UNESCO’s Ethics of AI~\cite{UNESCO}, Microsoft's Guidelines for Human-AI Interaction~\cite{amershi2019guidelines}, Google's People + AI Guidebook for designing human-centered AI products~\cite{google_pair}, NIST’s AI Risk Management Framework~\cite{nist2023airmf}, and the accompanying Generative AI (GenAI) Profile~\cite{autio2024genaiprofile}, which provides specific risk management guidance to address GenAI models and uses. UC Berkeley has also published an AI Risk-Management Standards Profile for General Purposes AI Systems (GPAIS) and Foundation Models~\cite{barrett2023ai}. While these guidelines recommend regular audits and systematic reviews and risk assessments of the models, they don’t provide recommendations on how AI models or AI-enabled products and services could continuously monitor the performance of the models by leveraging users’ direct feedback to detect issues. Such feedback allows for more timely and scaled signals, compared to risk assessments, audits, and reviews, which cannot be done on a continuous basis due to the complexity, cost, and time-intensiveness of the processes. 

Some of the existing frameworks and industry guidelines, such as Microsoft's Guidelines for Human-AI Interaction, explicitly call for systems to encourage granular feedback during and after the interaction (e.g., Guideline 15) 
\cite{amershi2019guidelines}. However, these guidelines are rather vague and do not provide operational guidelines about how to emplement effective feedback mechanisms. For example, they suggest ``consider[ing] how to convey to the user that the feedback was received'' or warn that it may be ``unclear to the user that they can provide feedback to the AI system'' or ``unclear to the user what the feedback options mean'', but do not provide actual design recommendations about what systems' signals, entry points, or feedback options convey such information to users clearly, and which approaches, in contrast, are ineffective or confusing. 
Similarly, one of the principles in Google's People + AI Guidebook for designing human-centered AI products~\cite{google_pair} recommends adapting AI using feedback to maximize product benefits and minimize potential harms to users. This Guidebook also emphasizes the importance of transparency about privacy-related implications of providing  feedback, including what information is being collected, when, what it is for, and how it can it benefit the user. While the Guidebook suggests designing structured multi-modal feedback collection mechanisms, it offers limited guidance about how to actually design feedback interfaces, with few suggestions about content.

\subsection{User Research about Feedback Mechanisms for GenAI Systems}
User-centered research about user feedback mechanisms in GenAI interfaces is scarce. 
One notable exception is a critical analysis of survey results about the feedback interface for ChatGPT, 
which indicates that the current design encourages simple, frequent, and performance-focused feedback, while actively discouraging collective input and discussions among users~\cite{cooper2025chatgpt}. Authors recommend exploring whether more participatory approaches and deliberative processes would better capture ethical and social concerns. 

A survey of 1,500 participants from 75 countries assessed their contextual model preferences and structured feedback regarding value alignment, fluency, factuality, safety, diversity, creativity, and helpfulness of over 8,000 conversations with 21 LLMs ~\cite{kirk2024prism}. It found that model rankings varied according to 
to idiosyncratic, contextual, and group-wise factors, and that average scores concealed model ratings and model preferences among minority groups, emphasizing the importance of diversity among human raters. Such idiosyncratic differences support the idea of developing personalized RLHF frameworks to tailor LLM responses to individual user preferences based on their feedback~\cite{li2402personalized,zollo2025personalllm,poddar2024personalizing} or inferred profiling~\cite{jiang2025know}. 
In an experiment, \citet{xu2023learning} collected different types of human feedback (including binary quality measurements, free-form text feedback, and fine-grained multi-choice reasons for model failure) to compare various methods for learning from feedback. They found that feedback about particular modules outperforms feedback about the overall output, and combining multiple modes of feedback results in best learning outcomes. However, neither ~\citet{kirk2024prism} nor \citet{xu2023learning} assessed the \textit{usability} of the feedback mechanisms themselves.


\subsection{Existing gap and our contribution}
To summarize, to the best of our knowledge, despite some companies mentioning the value of user feedback in their guidelines and some key principles~\cite[e.g.,][]{amershi2019guidelines,google_pair}, user-centric research evidence about effective feedback mechanisms for GenAI-enabled systems is scarce. There is currently no standardized framework, approach, or detailed design guidelines to capture high-quality, output-specific feedback in GenAI interfaces.
To reduce this gap, we systematically analyzed the current status quo across multiple interfaces, establishing an evidence-driven baseline, and conducted research with users and designers of generative AI features. This paper contributes concrete guidelines, an actionable framework, and design recommendations to advance the design of post-deployment feedback tools, transitioning them from ad-hoc interface elements into robust, human-centered governance mechanisms.

\section{Methods}
\label{sec: methods}

We conducted multi-phase research. To address \textbf{RQ1}, we started from developing a benchmark of existing approaches for collecting user feedback about GenAI across different sectors. Then we conducted usability inspection of those approaches, and developed a taxonomy of common design elements and heuristics for evaluating their usability (\S\ref{sec: benchmark}). To address \textbf{RQ2}, we invited eBay stakeholders to provide structured feedback via expert heuristic evaluations, to take into consideration cross-functional perspectives (\S\ref{sec: heuristic eval}). 
Based on insights from the usability inspection the heuristic evaluation, and 
input from eBay stakeholders, 
we developed an initial set of guidelines and best practices, which we used to design an initial prototype of the feedback tool. We then tested the prototype with eBay users (\S\ref{sec: user test}), to complement the researchers' and stakeholders' perspectives with users' opinions (\textbf{RQ2}). We used participants' feedback to refine the prototype and the guidelines.  

\subsection{Benchmark and Usability Inspection}
\label{sec: benchmark}
Using web and mobile traffic rankings, we identified the most popular digital products, 
including dedicated GenAI tools/platforms ~\cite{rank2024genai} as well as websites ~\cite{rank2025websites} and apps~\cite{rank2024apps,rank2025Androidapps,rank2025iOSapps} across major sectors and genres (e.g.\ search, entertainment, social media, news media, communication, productivity),
including e-commerce/shopping services. From those sets, we chose 
up to 5 products per category that offered GenAI features in their user interface and met our inclusion criteria, as well as the 7 dedicated GenAI tools out of the top 10 that met our inclusion criteria, and identified those that had user feedback mechanisms. We also supplemented with products that weren't on the most-popular lists to achieve better coverage across categories. 
When available, we included both the mobile and web version of the products. 

We excluded products that did not have a free version, required a device the researchers did not have (e.g., VR glasses or a smart TV),  
required a business account or significant commitment to set up a seller or advertiser account, 
raised security concerns (e.g., providers of adult content), were not available in English, or simply provided an alternative user interface to a GenAI product we already included (e.g., Facebook's portal to Meta AI). We included only products that were clearly labeled as using GenAI. 

The resulting list consisted of 32 
GenAI products that asked users for feedback, 
including 10 
GenAI tools (e.g., OpenAI ChatGPT, Google Gemini, Meta AI, Bing Copilot), 5 
GenAI features in e-commerce services (e.g., Amazon’s Rufus, Google Shopping’s AI Overview, Shopify’s Shop App AI), 11 
GenAI features in other websites and apps (e.g., Google Search, WhatsApp, Reddit), and 6 
eBay GenAI features (e.g., item description generator, AI shopping assistant); 
see Appx.\ \ref{appx:taxproducts} for a full list. 

We documented user interface and user experience (UX) design elements and navigation flows in the existing user feedback collection approaches for GenAI features in those products, and taxonomized the design patterns along 16 dimensions, including entry points, UX writing, visual elements (like icons), feedback format, etc. (see Appendix~\ref{appx:taxdimensions} for details). We also identified common usability issues and examples of good practices for collecting user feedback. We collected initial feedback on the taxonomy, solicited additional dimensions that would be useful to document, and discussed identified issues with 
cross-functional partners (including product managers (PMs), product and content designers, UX researchers, and legal and privacy experts) across different teams at eBay. 

\subsection{Heuristic Evaluation}
\label{sec: heuristic eval}

Based on 
Nielsen's heuristic evaluation method guidance~\cite{nielsen1990heuristic,nielsen1995conduct}, supplemented by 
AI-specific issues identified during the usability inspection (\S\ref{sec: benchmark}), we developed 11 usability heuristics (such as discoverability of entry point, ease of navigation, clarity of wording/information and process, consistency, etc.). Besides 4 eBay AI features available at the time, we selected 8 
popular 
products from the taxonomy, 
including 4 dedicated GenAI tools and 4 features in e-commerce services (see Appx.\ \ref{appx:taxproducts} for details). 
Then we asked 24 experts at eBay to systematically assess the approaches to user feedback collection in those products (including PMs, designers, UX researchers, legal, engineering, privacy, etc.). Each stakeholder evaluated 3 randomly assigned products (1 of each type). They independently evaluated how well each of the 11 heuristics was implemented on a 4-point Likert scale\footnote{We intentionally excluded the middle point, to clearly identify positive and negative usability practices.}, highlighted issues and positive aspects, and suggested improvements on each heuristic, using a provided template (see Tables \ref{tab:HE_template} and \ref{tab:heuristics} in Appx.\ \ref{app:HE_template}). 
Participants were identified among eBay stakeholders involved in designing or consulting on AI features (e.g., regarding legal advice), 
and invited either via direct emails, messages, or posts in relevant eBay Slack channels.  

We calculated the average Likert-scale ratings for each product per heuristic and overall, and 
synthesized common themes in the open-ended responses regarding issues, positive aspects, and recommendations.

\subsection{User Testing}
\label{sec: user test}
Based on insights from the studies described above and informal consultation with eBay stakeholders across a range of different roles about their use of feedback and future needs or plans for it, and in collaboration with relevant eBay teams, we developed an initial set of guidelines and designed an interactive Figma prototype of a user feedback tool. We mocked up entry points for 4 eBay GenAI features (2 for buyer-facing and 2 seller-facing ones) to make sure they were discoverable and relevant across different design environments, including text-, image-, and recommendation-based AI outputs. 

To evaluate the prototypes, we ran remote moderated usability testing with 13 US-based buyers and sellers on eBay. 
We distributed a screening survey to an eBay user panel, selected a mix of qualified participants (age 18+, native/bilingual or fully proficient in English, who passed the screening survey attention checks), and invited them to one-hour Zoom interviews. 
We took detailed notes during the interviews, which were supplemented by video- and screen-recordings and automatic transcripts.

Each participant was randomly assigned to one of the 4 eBay GenAI features and one of 10 scenarios describing potential issues they could encounter with those features across various categories (such as the system generating a biased, inaccurate, or offensive response), to assess whether the tool can adequately capture different issues users could face. When there was time, we tested additional issue scenarios and/or features, to collect more observations and to identify within-subject variations in feature- and issue-specific preferences.\footnote{The full user testing protocol is 
included in the Supplementary Materials.}

After the warm up questions, we showed participants a short video clip demonstrating a typical interaction with the eBay AI feature they were assigned, and asked them to imagine a scenario in which they encountered an issue while using that feature. Then we asked them to open the interactive Figma prototype, share their screen, and show us how they would inform eBay about the issue, while speaking out loud everything that was going through their minds, following the cognitive walkthrough method. We observed their interactions with the prototype and took notes about their actions, reactions, comments, questions, concerns, and confusions. If there was time, we also asked them to show us how they would inform eBay if they had a positive experience with the feature instead, without any issues. 

After they used the prototype tool, we shared our own screens and showed them one screenshot from the flow at a time, asking about clarity and comprehension, concerns, visual appeal, and ease of use, referencing the notes from the observation part when needed. For some of the screens, we also showed several variations in design and wording, to understand preferences. We updated some of the mock-ups and wording between the sessions, based on participants' feedback, to iteratively test these potential improvements.\footnote{The slides we used in this part are included in the Supplementary Materials. 
Most of the slides were used with all participants assigned a given example feature, but a few of the A/B variants were added during the testing.} 

Using deductive thematic analysis, we identified, discussed, and resolved disagreements about the common themes, challenges, and recommendations across different topics (e.g., regarding discovery of the entry point, clarity of wording, opinions about feedback modalities, doubts and concerns, and impact of the feedback experience on perceptions of eBay as a company). Based on the 
results, we refined the feedback tool prototypes and design guidelines (see the refined final version in Appendix~\ref{appx:guidelines}).

\subsection{Ethics and Confidentiality}
The study involving user participation 
determined to be exempt from IRB oversight 
by ICSI's Human Research Protections Committee (protocol 2025-025). Participants provided consent electronically on the first page of the screening survey. 
Heuristic evaluation was determined not to be human subjects research by ICSI's HRPC (protocol 2025-014). 

We associated users' responses with a pseudonymous participant number rather than their personal identifiers. 
We stored names and contact information for user testing participants separately from their interview data, and destroyed that information after recruitment and payments were complete. We also manually removed personal identifiers from the transcripts. 
We stored the recordings and transcripts in a password-protected 
folder that only researchers directly involved in this project had access to. 


\subsection{Limitations}
\label{sec:limitations}

The products included in our usability inspection and heuristic evaluations are limited to free popular services available in English, from a subset of industries. However, while the list is not comprehensive, it includes many commonly used digital products with GenAI features, with a large number of Internet users in the US, and often wide global adoption. Many of these products are developed by large companies that have more design, legal, research/analysis, and engineering resources than less popular products developed by smaller companies or individuals, and thus represent the best effort and leading practices in collecting user feedback about GenAI features. Since our goal was to develop design guidelines based on the identified common design patterns and usability issues (rather than to evaluate their prevalence), we believe that our selection of reviewed products was sufficient and effective in achieving that goal. Future work can explore design patterns and usability issues for GenAI feedback tools in other products, industries, and markets, and regularly monitor changes in a quickly-evolving AI landscape. 

This project was a collaboration between academic researchers and co-authors at eBay. 
This partnership allowed 
us to ground our findings and recommendations in a practical use case that accounted for how GenAI product teams currently use and want to use feedback, but necessarily incurred some trade-offs against breadth. Our expert heuristic evaluators were eBay stakeholders, whose opinions may be influenced by the company's culture and policies and by e-commerce-specific considerations, and therefore not representative of the opinions of all product teams. However, our procedures are generalizable and can be replicated broadly across companies and product types. 
Finally, our user testing participants were all eBay users, which supports high ecological validity, but naturally limits generalizability. Future work can engage broader groups of consumers as well as stakeholders from other companies and industries to identify additional nuances or differences between them. 

Due to the partnership with eBay, the design of the prototypes for user testing largely followed eBay's design and content guidelines, 
and the example features and scenarios were e-commerce-specific. However, the major design patterns (such as entry points, feedback formats, interactions, and navigation flows) are broadly relevant, and the \textit{types of issues} explored in the scenarios were fairly generic. In addition, where possible, we tested variations that went beyond eBay's design guidelines and technical limitations. 
Thus, our recommendations are relatively generalizable beyond the eBay or e-commerce scope.

\section{Results by Research Phase}
\label{sec: results}

Since many of the findings from the benchmark taxonomy/usability inspection, heuristic evaluation, and user testing overlap, we first present a brief overview of study-specific results for each phase in \S\ref{sec: results}, and then summarize the findings about usability issues that were informed by all those studies in \S\ref{res: usability issues}, to avoid repetition.

\subsection{Benchmark and Usability Inspection}
Out of the 40 
products with GenAI features we inspected, 8 did not have user feedback mechanisms at all. 
This suggests that collecting user feedback about AI features is an emerging industry norm, at least among industry leaders, with 80\% of the popular products in our sample already doing it. However, current implementations are often slight modifications of generic feedback mechanisms on overall satisfaction with the product/service or for reporting bugs, ignoring unique issues specific to GenAI, such as biased, offensive, or inaccurate responses or unrealistic generated images. 

In the reviewed examples, negative feedback was almost always requested, but positive feedback was solicited much less often, or in less detail. Although many of the reviewed mechanisms allowed users to submit feedback about the specific text response/output, feedback for images or the entire chat was much rarer.
We also identified common usability issues during our inspection, especially with respect to entry points, feedback modality options and flow, and the submission step (see \S\ref{res: usability issues}).

\subsection{Heuristic Evaluation}

The eBay team members who participated in the heuristic evaluation noted many of the same high-level issues as the researchers found in the initial usability inspection (focusing on ease of navigation, clarity about goals, and user value and encouragement). However, their varying expertise (across design, UX, compliance, etc.) allowed them to give in-depth detail on a breadth of specific issues with the evaluated feedback tools, such as accessibility blocks, unintuitive menu structures, gaps in privacy disclaimers, and hard-to-notice visual cues. They also identified good practices that could easily be adapted to other products/companies vs.\ good practices that were likely to be relevant only for particular use cases. 
In terms of ratings, we noted similar patterns and range in quality assessments for feedback tools for dedicated GenAI tools vs.\ features in e-commerce); neither domain led in usability or overall satisfaction.

\subsection{User Testing}
We interviewed 13 participants total, including 8 who were mainly buyers and 5 who were mainly sellers. Participants' ages varied between 36 and 70 years old. 
Nine participants were male and four were female. In terms of eBay tenure, 54\% had been using eBay for over 15 years, 23\% had been using it for 4--15 years, and another 23\% for 1--3 years. 
Only 8\% were using eBay AI features daily, 31\% weekly, 15\% monthly, 8\% at least once a year, and 38\% less than once a year. 
Each participant received 50 USD for completing the study. 

Some participants immediately recognized the value of user feedback in fine-tuning AI models and fixing problems. Others also saw obvious value in their feedback, but had doubts that product teams would actually read it or fix the reported issues.
After testing the prototype, participants said the fact that the tool seemed tailored specifically to GenAI features increased their confidence that the company paid attention to user opinions and would use the feedback to try to resolve issues. A few participants specifically noted that it is good to see platforms 
really trying to get it right when implementing innovative GenAI features. 

\section{Findings about Usability of Feedback Mechanisms}
\label{res: usability issues}
The insights below 
combine analyses of data from all our studies. 
They are structured roughly according to the 
dimensions that we derived from the usability inspection and used in the heuristic evaluation, though a few heuristics are combined where findings are closely connected. 
We refer to user testing participants (among eBay users) as \textit{participants}, and heuristic evaluation participants (among eBay stakeholders) as \textit{evaluators}. 

\subsection{User Value and Motivation}
\label{sec:uservalue}

The text of entry point labels and feedback form headers provides an opportunity to invite and encourage, but evaluators described the tone of some forms' text as ``transactional'', ``corporate'', and unfriendly or flat. There often was no mention of how users would directly benefit from providing feedback, or the user value was unclear and vague. For instance, ``your feedback will help improve the model'' does not emphasize a direct benefit for the users, but rather for the company. 
Similarly, explicit appreciation was rarely shown for users’ effort and contribution, e.g., via a ``Thank you'' message (some features had only a ``feedback submitted'' message or even just fading check mark icon being displayed). 
Uncareful wording may imply that only certain feedback is welcome (e.g., ``provide helpful feedback''), which can be off-putting for users. 

User testing participants appreciated seeing a thank-you snackbar (a brief temporary message that appeared near the bottom of the screen, 
in this case including a link to unsend the feedback) as an extra confirmation that their feedback had been submitted, and most liked the tone. Several participants said it gave them the impression their feedback would be taken seriously, though one suggested saying so explicitly, and explaining whether users will be contacted for follow-up. 

\subsection{Consistency}
There were frequent inconsistencies between mobile and web versions in content and/or flow that were not justified by the 
smaller screen or touch-screen interactions. For instance, the entry points or rating scales, 
multiple-choice categories (or their order) or feedback formats in general sometimes differed between the two versions. In some cases, only negative feedback was solicited in the mobile flow, but both negative and positive feedback was solicited in the web version; sometimes, there was no way to submit feedback in the mobile version at all, even though it was possible in the web version. Entry points sometimes had different text for labels or links, or used text labels in one version but not the other; 
wording and visual elements were inconsistent between web and mobile, or (where relevant) inconsistent with other similar feedback mechanisms on the platform or how features were named elsewhere on the platform, impeding clarity. Wording, shapes, or availability of buttons, and link and button treatment, were different between mobile and web versions, even when it did not appear to be necessary for screen size optimization. ``Thank you'' messages were not always present in both web and mobile versions. Privacy disclaimers were not provided in the mobile versions of some products, or gave different information between the mobile and web versions. 

\subsection{Entry Point}
\subsubsection{Entry Point Signal}
Thumbs up/down is a common, 
familiar, eye-catching signal for whether the AI feature gave a helpful/relevant response or not, and were easily chosen by participants given scenarios about unhelpful or irrelevant responses. However, thumbs up/down were not as obviously the correct path for scenarios that participants viewed as more serious (e.g., potentially harmful), or that they viewed as technical issues. Some participants given such scenarios initially passed over the thumbs to look for other paths---especially if they assumed the thumbs were only a voting mechanism, and not an entry point for detailed feedback. In addition to looking for reporting options or customer support links in menus, some participants expected (or at least wanted) to be able to report such issues via that chatbot directly, which they assumed would allow for a more in-depth back-and-forth. 

In some of the products assessed via heuristic evaluation, reporting legal issues was indeed a separate flow with a separate entry point from reporting other problems. 
Some menus included entry points for multiple related actions, like ``Send Feedback'' and ``Report Violation'' (or, worse, just ``Report'' without specifying what it was for). 

When we tested alternative entry-point signals, 
they were mostly dispreferred to thumbs. Specifically, a text hyperlink without thumbs did not allow for a quick, low-effort vote and could be less noticeable; a five-star scale seemed unclear in scope and overly nuanced (\textit{``too many shades of gray''}, P4), thus creating unnecessary cognitive load; a flag added next to the thumbs confused some participants as to how it differed from thumbs-down. 
Evaluators also suggested that five-level star ratings might feel over-precise for the nature of the question. 

\subsubsection{Entry Point Placement}
Entry points for providing feedback are not always intuitive or easy to notice. 
Specifically, they are harder to discover when 
1) users need to hover over the output in the web version or long-press on the output in the mobile version; 
2) they are not visually adjacent to the generated output, or at least visible on the same screen; or 
3) they are part of a menu rather than an immediately visible icon, 
making the flow longer 
and less intuitive 
(e.g., users might expect to find Settings or Chat History in a menu, but not feedback paths).
However, some participants with serious-harm or technical scenarios looked in the kebab menu for Reporting (or escalation) options. 

All participants noticed the thumbs up/down icons when they were positioned \textit{below} the AI output, even if they did not recognize their purpose fully. They parsed thumbs below the response more quickly than thumbs above, as they processed the screen top to bottom. 
When thumbs up/down were superimposed on an image (without any label), some participants noticed them quickly, but others did not notice them at all, due to low color contrast and the icons blending with some images. 

The placement of the entry point can help users understand what the feedback should be about, but does not solve all confusion. 
Even for apparently simple features (with a single generated item visible above the fold), participants disagreed on whether feedback was supposed to be about the just-generated content or the feature as a whole. Thumbs superimposed on a generated image were more likely (but not guaranteed) to be interpreted as being about the quality of that specific image. For complex features that produced several outputs (e.g., generated text 
and recommended listings), both evaluators and participants had quite varied interpretations of what an entry point below the whole response was requesting feedback about---the recommended listings only, listings and generated content, or the feature as a whole. 
However, the evaluators thought that providing separate entry points and flows for different types of content within a response could still confuse users as to which one is for which output, and would require more time/effort from the user if they wanted to provide feedback on multiple aspects or the whole feature. 


\subsubsection{Text Accompanying Entry Point} 
Text inviting feedback or providing labels for the entry point icons was often lacking entirely (with only the thumbs icon being present). 
Some evaluators appreciated the simplicity, and were concerned that text displayed next to entry-point icons could take too much space. 
However, a text label (especially if bolded) made the entry point more noticeable to the user testing participants. 

Where entry points did have text labels, some were vague (e.g., ``Report'', ``Downvote'', 
``How'd we do?''), 
were unclear about what the user is expected to report, rate, or provide feedback on, and/or missed the opportunity to encourage feedback or highlight user benefits from doing so (see \S\ref{sec:uservalue}). 
For many participants, ``Give feedback'' was sufficient to indicate this was an entry point for details, not just a vote. 
For complex features that produced multi-modal outputs, some participants thought the label could be especially useful in clarifying which part of the response the feedback was for. 

\subsection{Navigation}
\label{sec:nav}
Overall, navigation flows on mobile were typically longer, for example when long-press was the entry point instead of an icon (i.e., long-press > menu choice > feedback form instead of thumbs up/down icon > feedback form). Flows for reporting legal issues usually took more time/effort than providing feedback about other issues, due to having more steps and branches, 
including far too much text to process, and/or including information that was more clearly relevant to other products from the company. 

Even for non-legal flows, evaluators found some questions or feedback forms too lengthy, with too many different fonts/sizes, or unnecessary headings and subtitles. In a few cases, even buttons for basic actions like Submit were hard to find. 

\subsubsection{Feedback Formats}
Feedback formats typically include a binary vote (with thumbs up and down being the most common), and often (but not always) accompanied by a request for details via multiple-choice categories (pills/chips or check boxes) and/or open text comments. Some forms had several free-text questions (and no multiple-choice options), making the form look long, which could be discouraging for users. 

Not all participants and evaluators were expecting to be asked for detailed feedback after clicking on the thumbs icon (especially the thumbs up icon, as they viewed such details as less important than details about negative feedback; on the other hand, some appreciated being asked for details in either case). 
However, most participants appreciated having a set of quick preset options listing common and important issues, or positive feedback categories. Even those that preferred to use only the text box acknowledged that presets would make it easier for more people to leave feedback. Participants mentioned that having presets made it seem more likely the feedback would be used constructively, since it could more easily be routed to the right people working for the company that provided the AI feature. Similarly, participants appreciated having an option to provide free-text comments, and several believed feedback was more likely to be used if they both picked presets (for sorting) and wrote details in the text box (for specificity). 

Some evaluators were surprised that their response was sent immediately on clicking one of the preset options, with no opportunity to confirm. 

\subsubsection{Structure and Presentation of Multiple-Choice Preset Categories}
In some cases, multiple-choice lists allowed only one choice, which may not be appropriate if users want to report several issues. 
In the tested design using pills/chips---rounded capsule-shaped UI elements used for selectable items---the majority of participants either were not sure they could pick multiple presets or assumed (incorrectly) that they could not. Shown an alternative design with checkboxes, more participants recognized they could pick multiple options, but several were still unsure. Moreover, some participants preferred pills to checkboxes because of the larger target area. 

Too many preset multiple-choice options for negative feedback could be overwhelming. 
However, for the most part, participants viewed the list of eight presets we tested as being a reasonable length to balance the need to ``cover the bases'' against the need to keep it manageable for users. Several participants suggested adding ``Other'' at the bottom of the list, to make it clear that feedback on other topics was welcome.

Some evaluators were concerned that pills/chips 
might not work well for longer keywords. Evaluators also thought that preset options that included ``Other'' or ``Something else'' without a free-text box to specify further could feel uninviting (even for users who would not have picked ``Other'' anyway), suggesting that the company was only interested in predictable responses. 

\subsubsection{Responsible AI}
While multiple-choice categories always included model performance and output quality issues (e.g., relevance, accuracy, and tone), they only sometimes covered responsible AI (RAI) issues, such as harm, bias, or offensive content. When covered, they were not always included among the top-level options, resulting in lower discoverability and requiring users to exert more effort. 
Without explicit options, users may not be aware what RAI concerns can be associated with GenAI features, and that the company is interested in getting signals from users about those concerns. (This is also one of the disadvantages of not having preset options at all.) Some products included RAI-related concerns in their preset options, but had only one or two relevant categories, which may lead to some concerns being under-reported. Some products at least had a ``Report'' option, but not in the same pathway/mechanism where other types of feedback were submitted, 
so users might not have known to look for them. 

\subsubsection{Presentation of Free-Text Box}
The field label ``Do you have any other feedback? (Optional)'' led some participants to think the free-text box was specifically for issues not covered by the preset options. It should be clear that free-text comments can be used to add details about the issue indicated by the preset(s) or describe other issues. 

A few participants and evaluators found versions of the design where the free-text box seemed small (two or three rows tall) less inviting and suggesting the expectation of a short answer. 

\subsubsection{Option to Include a Screenshot or Recording Demonstrating the Issue}
When asked, most participants thought it could be useful to include an option for capturing and submitting screenshots or screen recording to document the issue, or at least were not opposed to it. However, when asked whether they would bother to use such a thing, many said they probably would not use it under most circumstances, especially as some assumed that session records would provide all the necessary context. However, if product teams find the context information currently collected alongside feedback insufficient, participants had several suggestions for how to make a screenshot option most useful/usable. Several participants wanted to be able to edit or annotate screenshots to clarify the issue. While screen recordings could provide richer context, most participants expected screenshots to be easier both to capture and to analyze. All participants said they would prefer if the screenshot could be captured automatically, with several saying they would be less likely to do it if they had to attempt backing out of the form, finding or replicating where they had the issue, and capturing the screenshot themselves. 

\subsection{User Control and Freedom}
Some feedback forms were not optional or had non-optional parts, or it was hard to tell which parts of the detailed feedback request were mandatory vs.\ optional. For instance, in some cases, the detailed open-text comments were non-optional (or design choices make it look mandatory, e.g., when 
there was no explicit label saying it was optional). 
In other cases, the multiple-choice question was non-optional, which might not be appropriate if the issue the user experienced did not fall into any of the provided categories. 

In the tested prototype, most participants understood that they could click ``Send'' to submit just their thumb vote without choosing preset options or leaving free-text comments, as those fields were clearly labeled as ``optional''. However, several participants wondered about the value to eBay of having a feedback form where nothing was required, or even suggested making the presets mandatory: \textit{``I don't know if the word ‘optional’ is necessary. Sometimes the word optional means you're off the hook, you don't have to go any further than this. ... but, you know, your intent is to get all the feedback you can''} - P11.
In contrast, a couple of participants commented positively on the optionality of providing details. 

In many examples, users could not edit or unsubmit feedback submitted by mistake. Sometimes it was not possible to submit feedback on multiple issues for the same interaction (e.g., by choosing more than one preset option or by submitting multiple feedback forms about the same results/interaction). (However, some evaluators and user testing participants found it confusing when they thought they \textit{could} submit multiple forms about the same results/interaction; see \S\ref{sec:status}.) 

In user testing, many participants did not notice there was an ``Unsend'' link in the thank-you snackbar until we drew their attention to it. However, several participants thought there were situations where it could be useful to be able to unsend or edit feedback (for example, if they sent something ``in the heat of the moment'', P10), and noted they would have been more likely to notice ``Unsend'' if they had been in such a situation. 
While 6 seconds was generally perceived as sufficient time to make a decision about unsending, some participants would have preferred more time (e.g., 10 seconds).\query{Cut for space? -AF}
Several participants wanted to have a confirmation step before unsending their feedback. 

\subsection{Visibility of System Status}
\label{sec:status}
Sometimes there was no indicator at all (or no persistent indicator) that feedback was successfully submitted (such as the thumbs icon changing color to filled), making it unclear to users whether their feedback was received or if there was a technical issue. Sometimes that indicator was too subtle or too far away from the last user action on the screen, so it was less likely to be noticed. Status signals may be too large/wordy and thus may feel too interruptive (for ephemeral signals) or too cluttered (for persistent text signals). 
In some cases, clicking on the filled thumbs again would change color to unfilled, but it was not explicitly explained whether that means the feedback got unsent/destroyed. 
Usually there was no way to see details of feedback that was already submitted. 

All participants understood that a filled thumb icon means that the feedback has been submitted. A majority of participants did not expect to be able to submit feedback multiple times for the same AI output using the same thumb icon. Therefore, many did not expect anything to happen if they clicked on the filled thumb again. However, a few participants assumed that clicking on the filled icon would open a new form (creating confusion about what happened to the previously submitted feedback), or that it might allow them to view or edit the submitted form. Asked what it would mean if the filled thumb reverted to unfilled when clicked, participants were confused; some said they’d assume it was a glitch, and some would worry that their submitted feedback was not recorded. Some participants thought the feedback tool might allow them to send both positive and negative feedback for the same output, but several opined that this would send mixed signals, and thus would not be helpful for eBay, or said it could create confusion about which actions would cancel their prior vote.
Most participants interpreted the thumb changing back to unfilled after the user clicked ``Unsend'' link in the thank-you snackbar as confirmation that it was indeed unsent.  

\subsection{Clarity of Wording/Information}
In addition to the header ``Share your thoughts'', most participants preferred a design where the label above the presets mentioned the specific feature the feedback is supposed to be about (e.g., ``Tell us more about the issue with eBay’s AI Shopping Assistant''). They found it more concrete, streamlined, clear, and inviting. 

Evaluators noted that some examples referred to features using company-internal names that users may not recognize (e.g., referring to an AI chatbot as ``the agent'' or ``the assistant'' without such names being used outside of the feedback flow). They also found that unclear legalese jargon those used in legal-reporting flows and in privacy disclaimers, potentially reducing user comprehension and making such disclaimers more alarming than helpful. 

\paragraph{Clarity of Preset Options}
The wording of the multiple-choice categories was often unclear (e.g., not explaining what does it mean for the model output ``Being lazy''), or contained jargon and abbreviations (e.g., ``text rendering'', ``overactive refusal'', ``bias'', ``UI bug''). 
Evaluators found it hard to differentiate between ``irrelevant'' and ``inaccurate'', or thought that some presets sounded over-technical (framed from the product team's point of view rather than the user's). They suggested balancing conciseness against clarity. 

Most participants found the wording of most presets in tested prototype versions clear, but they were sometimes uncertain about what the use cases were for a given category.
For the main flow, we used ``Inaccurate or unrealistic'' to cover a variety of problems with text and images. When offered alternative versions customized for specific features, many participants found “misleading” to be more clearly broadly applicable, and more important than “unrealistic”, especially for non-image features. 

We tested several presets to cover Responsible AI concerns, with varying success. Some participants were not certain how “Stereotyping or bias” would apply to GenAI features on eBay. Participants who chose it also picked “Offensive or inappropriate” alongside. Thus, while capturing concerns about bias separately may be beneficial for the purposes of ensuring compliance with responsible AI guidelines, it is likely that such issues could also be captured under a more general category like “Offensive or inappropriate”. 
“Unsafe” was not well understood, and participants seemed to primarily think of physical safety risks, rather than online safety threats like harassment or bullying. “Harmful or unsafe” was also paired with “Offensive or inappropriate” rather than being chosen by itself. 
We therefore recommend recombining these three presets into two that include the most operative terms, “Offensive or biased” and “Inappropriate or harmful”. 

Participants (especially those who were not given legal/copyright or privacy scenarios) had hard time imagining situations where the “Legal or copyright issue” and “Privacy issue” preset options would be relevant. They also doubted they (or other users) would be able to judge whether a situation is a legal, copyright, or privacy violation, so they had low confidence in using those presets. 
We recommend combining the two categories into “Potential legal, copyright, or privacy issue”, where “potential” emphasizes that eBay does not expect users to have legal expertise. 
Some product teams may prefer to capture such issues via a separate reporting flow for legal issues, when users pick this option, they could be re-directed to the relevant reporting page, or encouraged to submit the issue via both the present general feedback form and the reporting flow (e.g., under the kebab menu > “Help and Report” on mobile (if available), or using the “?” icon in the web version). 

Several participants viewed the high-level “Not what I was looking for” as vague, and viewed a thumbs-down as generally sufficient by itself to indicate that the offered options don’t appeal (i.e. they wouldn’t necessarily bother to select a preset about it). Participants were more interested in presets we presented as alternatives, especially for the seller-side features, that covered more specific problems like ignoring elements of the query or instruction, such as “Missing important elements” or “Left out important info”. 

Nearly all of the participants preferred the more concise versions in the tested prototype to alternatives presets in a more conversational format (e.g., “There was a technical problem” rather than “Technical problem”). Participants found the conversational versions unnecessarily verbose, more difficult to distinguish, and more work to read and comprehend (“\textit{The original is clearer because there are less words}” - P4), discouraging them from providing feedback. 

For the main testing flow, we presented the same list of presets for all four features. When we presented alternative versions of the form where a few of the presets were customized to the most likely concerns for those specific features (e.g., “Left out important info” for the Description Generator), a majority of those asked found the customized list more relevant to the feature in question, and sometimes found it easier to see what the categories could apply to. However, even if they liked the customized lists better, participants generally found the cross-feature list to be sufficiently relevant and reasonably comprehensive. 
Some product teams may prefer to use custom lists, depending on their priorities and capacity to design, parse, and analyze feedback from customized forms. 

The tested prototype did not include preset options in the positive feedback form, but about half of the participants thought presets could be helpful to speed up and simplify the positive feedback process. Participants suggested items such as ``good recommendations/listings'', ``accurate and relevant text'', ``liked the tone'', ``visually attractive'', and ``easy to use''. 

\subsection{Clarity about the Process, Including 
Additional Information, Privacy Disclaimers, and Transparency}
\label{results_clarityofprocess}


While in most cases, users were asked to confirm feedback submission, sometimes it was immediately submitted without confirmation or asking for details. This was more common for the positive feedback, since many features ask for details only where negative feedback is given. A few participants also mentioned that they would appreciate links to customer service or FAQs somewhere in the feedback process, to escalate or get help in troubleshooting the issues. 

In most of the products included in the usability inspection and heuristic evaluation, there was no indication of what happens once the user provides feedback, how it will be used (e.g., for training or fine-tuning model outputs for all users or for updating recommendations and preferences for the specific user based on their own feedback), who will see it (e.g., other sellers or buyers, third-party companies, companies that built the LLMs powering the AI features) and under what circumstances, 
what user data will be submitted along with the feedback (e.g., user ID, metadata about the device, browser, session, last prompt only or the entire chat), and what are the potential consequences of providing feedback (e.g., law enforcement involvement in case of serious violations or negative ratings on the platform; both of these expectation could be discouraging from submitting feedback if users find those measures too harsh of a ``punishment''). 

In some cases, there was a disclaimer in the feedback form outlining some of this information (but rarely comprehensively and clearly) and/or providing a ``Learn more'' link. 
In those examples, the disclaimer was not given upfront---information about what would happen to the feedback was usually shown only after the user had clicked on the thumbs up/down icon and brought up the detailed form, so in some cases the feedback vote had been already sent by the time the user saw the disclaimer, without asking for confirmation. Disclaimers were not always placed near the form's Submit button, potentially hindering awareness. 

``Learn more'' links usually went to general disclosures for the whole product, which rarely explained what happens with feedback specifically. The landing pages for ``Learn more'' were often long, or had multiple nested pages and hyperlinks, making it hard to navigate and find specific information. 
Some were policy documents rather than more user-friendly information formats, where 
the difficulty of finding specific information may discourage users from providing potentially useful feedback.
Moreover, use of italics or color alone to indicate a hyperlink for ``Learn more'' may be not understood by users, and  contrast of the hyperlink color was low in some examples, violating the WCAGG contrast ratio guidelines~\cite{caldwell2008web}. 

All participants found the privacy disclaimer we tested (which is not the same as eBay's current disclaimer) to be generally clear, saying it had standard provisions and language---though some noted they would be unlikely to read it in real life. For the most part, reading the disclaimer did not change participants’ sentiments about 
feedback collection. Most also said it did not change their understanding about the 
data practices, but a few found it improved their understanding. However, some participants wanted the disclaimer (or the referenced links) to include specifics if anyone besides the company 
itself (e.g., other users, third parties, AI services) would have access to submitted user feedback. 
Even before reading the disclaimer, many user testing participants expected that the AI output and the interactions that led to the issue would be included alongside a thumbs-down vote and the details they put in the form. However, only two participants mentioned that additional data not entered by the user (such as metadata, location, or purchase history) might be included, or retrieved via user ID. One participant was displeased to see that a session record would be included, especially as it was unclear how far back that might stretch. However, most participants thought such session context could be useful for resolving the issue. 

When we showed an alternative version of the form where users could use a toggle to specify that their user ID should not be sent with their feedback, some participants recognized the privacy benefits. However, in practice, a majority of participants were not very concerned about non-anonymous feedback, and a few said they would be skeptical about whether the toggle actually worked (i.e. they expected their data to be sent along with their feedback regardless of their interaction with that toggle). Some participants actually viewed it as beneficial to include user IDs, to provide more context for resolving issues and let eBay contact them with follow-up questions. In fact, a few participants assumed the purpose of the toggle would be for them to request follow-up. 




\section{Design Recommendations}

Based on the detailed investigation described in the previous section, we developed a comprehensive set of recommendations for designing feedback tools for GenAI products and features. This section is organized according to the same heuristic-based dimensions as \S\ref{res: usability issues}. A concise 
set of guidelines ordered according to the feedback flow, including example screen mock-ups, can be found in \ref{appx:guidelines}.


\subsection{User Value and Motivation}
\label{recs_uservalue}
Use friendly, inviting, and conversational tone in the text of entry point labels and feedback form headers, highlighting what value will users (not just the company or platform) receive from providing the feedback, e.g., the model's ability to suggest better or more customized recommendations (rather than benefits like ``model improvements'' that do not clearly connect to user value). Positive tone can also be conveyed through visual elements, such as animating thumbs when clicked. 

After users submit the feedback, thank them for the effort of providing feedback, and reassure that although the company may be unlikely to follow up with users individually, their feedback will be taken seriously. This is another opportunity to highlight user value (however, if space is limited, and user value can only be highlighted once, prioritize doing it earlier in the flow, to encourage users to provide feedback). Future research can test alternative message formats, as snackbar functionality and word count is limited.

Transparency about data practices builds trust, and thus further supports motivation. Therefore, explain how feedback will be used (see more details in \S\ref{disc: privacy}).

\subsection{Consistency}
To avoid confusion, for a given product or feature, the same user feedback form should be offered across mobile and web (to the extent possible given different affordances), with the same modes for requesting feedback (ratings, multiple-choice presets, and free-text comments), the same preset multiple-choice options (ideally, in the same order), and the same disclaimers and ``Learn More'' info. 
Ensure consistency in wording (e.g., of thank-you messages and preset options), visual appearance (including entry-point icons), 
and link and button treatment across products, mobile and web experiences, and different screen sizes. 
Ensure consistency of text strings with system design, and content and brand guidelines. 
Teams should consult with content designers for up-to-date guidance.

\subsection{Entry Point}
\subsubsection{Entry Point Signal}
\label{recs_entrypointsignal}
We recommend using high-contrast thumbs up/down icons as the main entry point for feedback on all types of issues, as it imposes less cognitive load than a scale (e.g., star ratings).
If space is particularly limited, product teams may consider a single entry point for both positive and negative feedback, with an opportunity to select between the two votes, once the feedback form is open. 
We advise providing a ``Report a problem'' option as an additional separate path (e.g., under a kebab menu or other relevant menus
, with a flag icon), that can be used to report bugs, potentially harmful content, or legal concerns, in case users do not see thumbs as a likely (or ``serious enough'') entry point for feedback on those issues. If feasible, it is recommended to add a direct link to customer support and/or a troubleshooting FAQ somewhere in the feedback flow (e.g., in the thank-you message; see \S\ref{recs_uservalue}, 
to allow users to escalate the issues, hear back from the company, or troubleshoot more common issues). 
Ideally, entry points should be visible, without the user needing to hover over or long-press on anything. 

\subsubsection{Entry Point Placement}
For simple features that generate one content item (e.g., only text response, or only an image), position the feedback entry point immediately below the generated content, where users already expect to see action options, like Copy and Edit. The heuristic evaluators also liked the alternative placing of the entry point to the right of the response in the chat-style designs; product teams can test that placement if it is appropriate for the design of their product. 

For complex features that return multi-modal 
responses or multiple recommendations or listings, further testing of the optimal entry point placement may be needed. When developing such multi-modal prototypes, we suggest obtaining separate feedback streams for each part of a complex feature, 
placing separate entry points as close as possible to each part (e.g., icons superimposed on the content, with high contrast, following accessibility guidelines~\cite{caldwell2008web}), with labels. 
If multiple entry points are not feasible, consider allowing users to choose which part they want to provide feedback on. 

To solicit feedback on the whole feature and all the things it does, thumbs at the bottom (below all the results) offer the best clarity based on our user testing results. However, this may sacrifice the quantity of responses if users do not scroll all the way down, and it is incompatible with ``infinite scroll'' design features. 

\subsubsection{Text Accompanying Entry Point}
\label{recs_entrypointtext}
We recommend using a boldface text label above the thumb icons inviting users to provide feedback in a friendly tone. A concise phrase like ``Share feedback'' can be used for simple cases. For more complex features, it may be helpful to specify the scope of the requested feedback, if there is space (e.g., ``Share feedback on image'' vs. ``Share feedback on [feature name]''). 
Supplemental signals later in the flow (e.g., headers or drop-downs within the feedback form) can further clarify the exact scope of the feedback request. 
It may still be helpful to include preset multiple-choice options that pertain to all parts of the response in each feedback form, in case users click the wrong one or want to leave feedback on multiple aspects.

Additional hover-over text for the entry-point icons can be used in the web version (e.g., ``Give positive feedback'' / ``Give negative feedback''). Note that ``Like''/ ``Helpful'' and ``Dislike''/``Not helpful'' do not necessarily reflect the variety of feedback categories, and could prime users to focus on aesthetics or user value rather than other aspects of the output (such as RAI issues). 
After the initial release of the feedback mechanism, it can be helpful to also use a tooltip to increase user awareness and comprehension of it.  

\subsection{Navigation}

Follow accessibility guidelines~\cite{caldwell2008web} to make sure all content and actions are easily discoverable and usable. 

\subsubsection{Feedback Formats}
Besides the positive/negative feedback (such as thumbs voting), allow users to provide optional additional feedback via quick preset multiple-choice options and free-text response (regardless of whether users picked ``Other'' or any other preset category), as it helps to both systematize feedback, and signal intent to analyze and address feedback to the users. 
While implicit feedback mechanisms were out of scope in this research, future work can explore such mechanisms, e.g., by provide clickable follow-up question recommendations in the AI chatbot, to help users realize that they can obtain more information or refine the results in a conversational manner.

\subsubsection{Structure and Presentation of Multiple-Choice Preset Categories}
For both positive and negative feedback, include 
six to eight quick preset options for users to choose from. Include ``Other'' 
as the last option to make it clear to the user that their feedback is not limited to the provided preset options. If space is limited, consider a nested structure with top-level categories (e.g., by general type or source of problem) and more detailed subcategories. Such nested or drop-down menus take less space, but consider that they require extra user action and do not provide a quick overview of all the preset options until users click on them. 
Preset options for negative feedback should cover the issues users are most likely to encounter, as well as RAI concerns, and items of particular interest to the product team; see examples in the guidelines in Appx.\ \ref{appx:guidelines}. 
Product teams may wish to include customized lists reflecting their priorities instead of using the templates. However, using standardized templates (at least across different features within a company or a product) allow to compare user signals across features.    

We recommend monitoring the use of different preset options, then place often used ones on top of the list, and maybe remove or re-word rarely used ones. However, we do not recommend removing RAI categories, because they provide an important signal, despite potentially rare incidents/use. 

Allow users to select multiple preset options at once (and include ``Choose all that apply'' in the label for the preset options) instead of submitting several feedback forms to signal various issues. 
Use designs with relatively large target area to make it easy for user to click/tap on them (e.g., pills or chips---rounded capsule-shaped UI elements used for selectable items---
instead of checkboxes). 
Other design consideration to keep in mind: checkboxes may allow for longer text as they have less padding than pills/chips, but checkboxes require vertical alignment, which would fit well in the layout on mobile devices, but pills/chips do not require vertical alignment and can take less space, especially in landscape/horizontal orientations, which are more typical for web versions. 

\subsubsection{Responsible AI}
Include preset multiple-choice options reflecting Responsible AI principles, e.g., allow users to report offensive, harmful, biased, inappropriate, or unexpectedly sexual images, text, or product recommendations.  
Consider providing additional channels/flows for flagging, reporting, or escalating particularly serious issues (such as harmful or illegal content). 

\subsubsection{Presentation of Free-Text Box}
Provide a substantial free-text box with an unambiguously broad field label such as ``Do you have any additional comments? (Optional)''. Make the input field a medium size to encourage or allow for longer open-ended comments.

\subsubsection{Option to Include a Screenshot or Recording Demonstrating the Issue}
We do not recommend including an option to capture a screenshot or video of the potential issue, unless preset options, free text, and session records sent along with the feedback are not sufficient to clearly understand problems with a particular feature. If product teams find it useful, we suggest offering users an option to easily attach a screenshot that is captured automatically (without requiring users to close the form to try to replicate the issue, potentially losing the current state, or creating concerns about losing it), with a possibility to edit or annotate it. 

\subsection{User Control and Freedom}
Make exiting the feedback form without submitting it easy and intuitive, e.g., by using scrim, ``Cancel'' button at the bottom, and/or X or back button in the corners. 
Make the preset multiple-choice options and free-text responses optional, and make it clear they are optional (e.g., by labeling it as such in the header), in addition to making the ``Submit'' button immediately active, without forcing users to provide details in addition to their thumbs vote. 
Use headers that intuitively emphasize the optionality of providing details, e.g., ``Would you like to tell us more?'' or ``Can you tell us more?'' instead of ``What did you like/dislike about this response?''. 
(Note that if a user thinks they have to provide detail but they do not feel like it, they may just exit without even submitting the thumbs vote, and a binary signal without details is still better than no signal at all).

If functionality allows, allow users to unsubmit the feedback, at least within a short period of time (e.g., 5-10 seconds) after submitting. Users should be able to confirm that action before feedback is unsent/deleted. 
Future research can test an option to edit feedback (in addition to, or instead of unsending it). For example, a temporary ``Edit'' link/button could return users to the filled form and allow them to edit and ``Send'', or to ``Unsend'' it.

\subsection{Visibility of System Status}
Highlight selected preset multiple-choice options clearly within details dialogs. 
If there are nested preset options, make sure the higher-level categories/choices are still visible while the user is being asked to choose the lower-level categories (e.g., using drop-downs within the same dialog, or if generating a new dialog to make lower-level choices, use the higher-level category as a heading). 

Change the thumbs icons' colors from empty to filled to indicate that the feedback has been submitted.
In parallel, show a brief confirmation that feedback has been submitted, unsubmitted or updated after editing, e.g., with a 2-second concise pop-up message. 
Allow users to view the feedback they have sent (e.g., by hovering over 
the filled thumb).

Based on the user testing findings, the least confusing pattern is to allow only one feedback form submission per AI output, and disable interaction with the thumbs after feedback has been submitted (regardless of whether the snackbar or thank you message is still visible).

\subsection{Clarity of Wording/Information}
If space allows, use a subheader, label, or contextual prompt that calls out the specific feature or user action the feedback is meant to be about, in a readable font size.
Use concise, clear, user-friendly wording and keywords in preset options and other text. 
Avoid technical and legal jargon, e.g., instead of ``bias'', consider using easy-to-understand words, like 
``offensive'', ``unfair'', ``one-sided'', ``shows prejudice.'' 
Ideally, evaluate the wording comprehension in user testing. 
Even for issues where long passages of legalese text are necessary
, minimize information overload by only providing the information that pertains to the specific product or feature, and is concise and truly necessary, instead of referring to the entire privacy policy or other long document that contains information not relevant to the feature. 

\subsection{Clarity about the Process, Including 
Additional Information, Privacy Disclaimers, and Transparency}
\label{disc: privacy}


To implement progressive disclosure, and balance transparency with usability and space limitation, include links to further information (e.g., via ``Learn more'' links) in a brief privacy disclaimer, placed where it is relevant to decision-making, e.g., next to the ``Submit'' button. Specify that those links will say more about who can see what data (especially whether other users, third-party AI services, or law enforcement (e.g., in case of reported legal issues) will be able to access it), and how that data will be used (including if users can expect to be contacted by the company about their feedback, if there could be any legal repercussions to the user, other buyers or sellers, or the company, for reporting issues). 
%
If space is particularly limited, disclaimers within the feedback dialog could be collapsed, with an expansion icon (such as question mark ``?'' or exclamation mark ``!'' icons). However, note that such collapsed version may be less easily discoverable and clear to the users.  
While users rarely read small-text disclaimers, their presence alone instills confidence that one can find out more about it if they wanted to, and build trust towards the company. 
We recommend crafting a tailored landing page for the ``Learn more'' link that explains data practices with respect to user feedback (e.g., in the FAQ format), rather than linking to generic explanation of data practices for the whole site/product, privacy policy, privacy notice, or user agreement documents. 
Consider making this information always accessible (e.g., on a company's website) even after the feedback form is closed. 
Wording of disclosures should be carefully crafted to be clear and concise, while avoiding scaring users. 
Make sure that disclaimers about how user data is handled meet country-specific legal requirements. 
Use clear indicators for the ``Learn more'' links, in alignment with the company's design guidelines and WCAGG accessibility guidelines~\cite{caldwell2008web} (e.g., underline or underline plus another signal (such as different color or bolding). 

Although in our user testing, we did not find support for users' desire to anonymize their feedback, we encourage product teams to test that option with their users, as these preferences may be culture-, product-, and context-specific. 


If feasible, the thank-you message after users submit the feedback could link to a troubleshooting FAQ page, customer support contacts, and/or a path for \textit{reporting} certain types of issues.

\section{Discussion and Future Work}
\label{sec: disc}


\subsection{Implications: The Value of Usable, Actionable Feedback Mechanisms for GenAI}
\label{sec: implications}


For a long time, human participation in development of AI models was limited to human-generated content in training datasets, which led to biases, unfairness, and other ethical issues~\cite{denton2020bringing,sloane2022participation}. 
Much of the existing work on AI governance and participation still focuses on pre-deployment or active development stages, such as co-designing fairness checklists \cite{madaio2020co} or establishing operational Responsible AI (RAI) guidelines for diverse technical and non-technical stakeholders \cite{constantinides2024raiguidelines}. 
Post-deployment user feedback embeds another form of participation that can help detect biases, ethics breaches, and other issues.

However, post-deployment mechanisms currently in use often rely on binary thumbs up/down mechanisms (which can accumulate a large but shallow pool of feedback). 
Prior work on use of input for model tuning has shown that such binary feedback signals fail to capture the multi-dimensional nature of generative outputs ~\cite{collins2025beyond,wu2023fine}. 
As AI agents progress along a spectrum of autonomy, users' roles are shifting from that of active operators who instruct the system and evaluate outputs, to observers who monitor system behavior, within minimal involvement in system actions  \cite{feng2025levelsautonomyaiagents}. 
This increasing autonomy, and the accompanying complexity of those systems, make it more difficult to engage users meaningfully. In this context, binary feedback signals 
are insufficient to capture dimensions such as safety, bias, fairness, and representational harms that are highlighted as societal risks in human-centered AI research
\cite{tahaei2023hcerai}. 

At the same time, prior work has shown that richer feedback is difficult to elicit and use effectively~\cite{collins2025beyond}. 
We found that richer modes of explicit feedback collection, when used, are often not aligned with usability best practices, and are not structured in a way that is clearly useful for improving products. 
While user feedback systems are expected to function as data collection infrastructures for alignment with evolving human values, their usability limitations and weak integration into pipelines reduce their effectiveness. (Similar to the gap between high-level design principles, such as privacy by design, and practical implementation by developers \cite{tahaei2023embedding}.) This gap persists despite urgent calls to action, including recent mandates for post-market monitoring by NIST \cite{nist2026monitoring}, and broader calls to democratize AI inputs \cite{vincenzi2024lay,brown2019toward}. 

This suggests an opportunity to bridge post-deployment monitoring with real-world, user-driven feedback loops that more directly inform model updates and system behavior. 
However, research is sparse on how post-deployment feedback mechanisms \textit{should} be designed. 
Our research sought to address this gap in both research and practice, by providing practical, actionable guidelines for companies to advance beyond the current status quo.

By providing design recommendations, a flexible prototype that is relevant for a range of contexts, and a model for 
stakeholder consultation, 
we aim to help companies streamline their feedback processes for fine-tuning AI models, across product teams with a variety of goals relevant for those companies. Our engagement with an industry partner, eBay, helped us to ground our recommendations in practical needs and considerations, and helped eBay to optimize their feedback processes and be better prepared to implement requirements for developing responsible AI and protecting users from harm.

At the same time, our results contribute to human-centered AI by investigating how feedback is given and used in GenAI products and features and what concerns are centered by current feedback mechanisms, 
positioning feedback mechanisms as socio-technical systems rather than purely interaction features.

\subsection{Future Work: Expanding User Engagement beyond Feedback Mechanisms}


Recent trends in critical AI scholarship position user feedback mechanisms as a \textit{first step} towards meaningful participation in AI development, with deeper engagement provided by participatory research that involves communities in decision-making and aims to democratize technology design \cite{huang2024collective,birhane2022power,bondi2021envisioning,crawford2021atlas,hashmati2025eXplainable}. 
User feedback tools in GenAI interfaces are a (relatively) feasible way to invite users to react to system outputs, but often do so in highly structured, provider-defined ways. They thus sit squarely in the tension between the need for practical, actionable input that meets product teams' goals, and the need for fuller participatory mechanisms that encourage product teams to align their work with societal goals.

The goal of the research described in this paper was to increase participation by providing guidance 
for making user feedback mechanisms a more usable, meaningful, and practicable part of AI development. However, developers and designers of AI systems should not limit users' participation to rating- or survey-based user feedback mechanisms only, and should explore other forms of meaningful engagement of users in the design process~\cite{delgado2023participatory,corbett2023power}.
Recent research suggests that ``participation brokers'', who facilitate the inclusion of users into shaping of AI systems, are well-positioned to scaffold such participation through user education, advocacy, speculative design activities, and developing methodologies for integrating user feedback into AI development and related policy processes~\cite{cooper2024fitting}. 

Participatory AI research frequently references Arnstein's foundational ``ladder of citizen participation'' \cite{arnstein1969ladder} to categorize participatory AI efforts. For example, \citet{sloane2022participation} categorize participation into \textit{work, consultation,} and \textit{justice}, while \citet{delgado2023participatory} propose a four-level framework: \textit{consult, include, collaborate,} and \textit{own}. Within these frameworks, the 
collection of user feedback in GenAI systems 
can be considered ``work'' or ``consultation.'' 
There is a need to explore how higher levels of participation can be made more feasible.

In particular, new post-deployment modes of participation should be explored. As \citet{feffer2023from} note in their literature survey, 
embedding structured participation into AI requires assessing how systems engage stakeholders throughout 
deployment. 
Moreover, some stakeholders in our study 
expressed a preference for feedback mechanisms that support ongoing 
engagement rather than one-off signals. They envisioned feedback as 
an iterative process through which users are consulted and involved in shaping systems' behaviors.  This aligns with broader work on iterative feedback loops and continuous learning in socio-technical systems \cite{thoughtworks2023responsible}.

As AI systems increase in autonomy, feedback mechanisms must be reconceptualized as end-to-end socio-technical processes embedded within AI governance. Rather than isolated UI features, they should enable transparency and accountability across multiple levels: informing model development/improvement and product decisions, supporting regulatory compliance, and, at the highest levels of participation, providing consumers 
with meaningful insight into how systems operate. 
Such systems could play a critical role in aligning increasingly autonomous AI agents with evolving 
human values at individual, organizational, and societal scales.

Standardization of user feedback mechanisms, on the one hand, facilitates building more efficient pipelines between user input and model fine-tuning, but on the other hand, standardization may result in homogenization of user input, creating risks for amplifying biases, suppressing diverse perspectives~\cite{eubanks2018automating}, and imposing values not congruent to specific contexts~\cite{gabriel2025matter}. Prior work has also emphasized the importance of accounting for individual characteristics of raters, annotators, and feedback contributors~\cite{goyal2022your,prabhakaran2021releasing}, including dempographic differences in how users interact with feedback systems \cite{mishra2026decodingsafetyfeedbackdiverse}. 
For example, the use of ``helpful, honest, harmless'' framework in RLHF~\cite{askell2021general,bai2022training} does not account for cultural and epistemic differences in the definitions and cross-context interpretations of those values~\cite{denton2020bringing,sambasivan2021re,gabriel2025matter,davani2024d3code}. 
Thus, it is important not to constrain user feedback to predefined multiple-choice categories only, but rather to allow open-ended comments and integrate other forms of user participation, to complement user feedback mechanisms. 

Challenges remain in capturing representative public perspectives 
and designing feedback systems that 
reflect diverse user experiences \cite{tahaei2024surveysharmful}. 
Looking forward, there is an opportunity to develop holistic frameworks that link individual-level feedback on specific outputs with aggregated, population-level insights, providing a longitudinal view of societal expectations.  
A survey of participatory AI projects found that they have been focused on a few homogeneous user groups~\cite{delgado2023participatory}. Scaled technical systems that do not account for diversity can exacerbate existing inequalities and further marginalize underrepresented views~\cite{eubanks2018automating,kirk2024prism}. Thus, future work can explore opportunities to broaden participation and better include underrepresented communities in the design of AI systems. For instance, our evaluations of current mechanisms noted some accessibility issues. Greater attention could be paid to ensuring accessibility in feedback mechanisms, including multi-modal forms of participation (e.g., not just just text-, but also voice-based feedback), to engage users with lower literacy levels or low vision. (Multi-modal feedback is also shown to result in superior model training~\cite{xu2023learning}.)
In addition, collecting metadata and socio-demographic information about users who provide feedback can help control for potential sample bias. The PRISM framework~\cite{kirk2024prism} could be utilized to further ensure the collected feedback is Participatory, Representative, Individualized, Subjective, and Multicultural.

\section{Conclusion}

In a collaborative study between academic researchers and eBay partners, we explored the landscape of how products with GenAI features currently collect feedback from users and heuristically evaluated example feedback mechanisms; conducted research 
with eBay users about how they view 
feedback processes; and developed and user-tested a flexible, evidence-grounded prototype feedback tool. In the process, we identified common issues with current tools and developed recommendations for requesting feedback from users of GenAI products and features. Major recommendations focus on making feedback entry points easily recognizable and inviting; using language that shows appreciation for feedback and motivates how users will benefit from giving it; giving the user options to provide simple binary feedback, quickly pick from among common issues, or provide details in text, letting them choose their level of effort and engagement; and providing clear signals about when feedback has been submitted and what will happen to it.

\section*{Acknowledgments}
\todo{Mohammad to get people's permission to be thanked by name.}\query{As of 7/7/2026, have permission from Circe, Imman, and Sujayendran.}
This research was supported by eBay's Research and University Partnership for Technology (eRUPT) grant program, 
and many individuals at eBay shared their time and expertise with us.
We especially thank Circe Granholm for 
insights into eBay's current user feedback processes and for structural suggestions; Barbara Kofler for contributions to the prototype and content guidelines; and Imman Abishek, Sujayendran Surendran, and the eBay Expressions team 
for assistance in recruiting users to test the prototype.
We are also grateful to the expert heuristic evaluators, and to other eBay stakeholders who shared their experiences and suggestions.

\section*{Disclosure of Interests}
\label{sec:disclosure}

The work of two of the authors (Frik and Bernd) was supported by a grant from eBay to the International Computer Science Institute, via its eRUPT program that promotes collaborations between eBay and academic researchers. The other two (Karami and Tahaei) are eBay employees, and worked on this research as part of their regular employment at eBay.\footnote{Tahaei is also affiliated with the International Computer Science Institute, but worked on this project in his eBay capacity.} As described in the paper, we used current eBay features as examples in some studies and consulted with various eBay employees across different teams, and the heuristic evaluators were from eBay. eBay reviewed the paper manuscript and made edits to avoid revealing confidential information in conflict with their interests.

eBay plans to implement some of the recommendations described in the paper, including using a version of the prototype tool as the basis for a new feedback survey template for GenAI features.  

Research design and conduct were led by the non-eBay researchers, who do not have any personal financial interests in eBay, to ensure the project focused on scientific investigation of the research questions. In particular, to avoid potential bias, user testing of the prototype was planned, conducted, and analyzed by the non-eBay researchers. We explained this to participants, to allay any concerns about giving offense and avoid any potential worries about consequences on the platform. eBay employees did not access documentation linking participants' eBay IDs and other PII with user testing responses.

For transparency, we mention the collaborative nature of the project in the abstract and Introduction to the paper (\S\ref{sec: intro}), and discuss both the advantages and the trade-offs of the project's focus on and engagement with eBay in the Limitations section (\S\ref{sec:limitations}).

\section*{Use of Generative AI}

We used Figma Make to assist in converting the human-composed screens in the prototype into a clickable live version for the user testing.\query{Amitis says: ``I don't remember the exact version, I'm not sure if Figma announced it at that level.'' // I tried to figure it out, but I couldn't even find a current version number for Figma, much less historicals for sub-products. -JB} The resulting flows were checked by a human to make sure they worked as intended, and that the original text and visual content were not altered.

Some of the suggestions in the visual guidelines (Appx.\ \ref{appx:guidelines}) were edited for smoothness using ChatGPT (various versions). The resulting text was checked by a human and, where necessary, revised to maintain the originally intended meaning.

Some passages in the paper were edited using Gemini 3.1 Pro and GPT 5.5 Pro to ensure the writing was clear, consistent, and grammatically correct. The human who wrote the original draft of those passages reviewed the proofreading/edits to make sure the original tone, content, and facts were preserved.




\section*{Data Availability}


We are not releasing user walkthrough/interview transcripts 
because participants did not consent to public sharing of their data (even in de-identified form). 
However, to support reproducibility, we have included 
the walkthrough and interview protocol in the Supplementary Materials.

\bibliographystyle{unsrtnat}
\bibliography{refs}  

\appendix

\section{Benchmark Taxonomy of GenAI Products and Features}

\subsection{Dimensions for Taxonomy and Usability Inspection}
\label{appx:taxdimensions}


\begin{itemize}
\item Entry point/When feedback is requested
\item Text inviting feedback at entry point (if any)\begin{itemize}
\item Including hover-over text \end{itemize}
\item What it is soliciting feedback about (the whole feature, or some of its parts)
\item Type of feedback (e.g.\ rating only (binary or scale), preset options, free text)
\item List of all preset options for negative feedback (if applicable) \begin{itemize}
\item Is Safety/Harm included in the options? If so, is it at the top level?
\item Is Bias included in the options? If so, is it at the top level?
\item Is Accuracy included in the options? If so, is it at the top level? \end{itemize}
\item Whether detailed feedback (presets and/or free-text) is optional
\item Confirmation or thank-you text after positive feedback is submitted (if any)
\item Confirmation or thank-you text after negative feedback is submitted (if any)
\item Persistent signal that feedback has been submitted (if any)
\item Whether feedback can be unsent and/or edited
\item Whether there is a link or pop-up to learn more about the feedback process
\item Whether a login is required to submit feedback 
\end{itemize}

We recorded screenshots of entry points, forms, and messages for each product to accompany the taxonomy. We also documented information about the product the feedback was solicited for, such as what it does (e.g., generates text/image/video, recommends products), whether it requires a login, and whether the user has a choice about interacting with the feature, as well as the name and sector of the company providing the product.

\subsection{GenAI Products Reviewed}
\label{appx:taxproducts}


The lists below include all products reviewed in the benchmark taxonomy. * indicates the products we selected for the Heuristic Evaluation.

Flagship GenAI tools:
\begin{itemize}
\item ChatGPT/DALL-E*
\item Claude* 
\item Duck.ai (anonymous portal to other tools)
\item Google Gemini*
\item Grok
\item Meta AI* (also used for WhatsApp search and Instagram search)
\item Microsoft Copilot
\item NightCafe
\item Perplexity
\item Poe
\end{itemize}

GenAI features in e-commerce products/services:
\begin{itemize}
\item Amazon Rufus*
\item BuySmart AI
\item Google Shopping AI Overview*
\item Google Shopping Try-On Models*
\item Shop App AI (for Shopify)*
\end{itemize}

GenAI features in products from other sectors:
\begin{itemize}
\item Bing Copilot Answers (search-based feature)
\item Bing search summaries (not part of Copilot Answers)
\item DuckDuckGo Assist
\item Google Search AI Overview
\item MacOS Preview Writing Tools (desktop app)
\item Reddit Answers
\item Snapchat My AI
\item WhatsApp Sticker Create
\item Yahoo News Takeaways
\item Yahoo Search summaries
\item Yandex/Dzen Search summaries 
\end{itemize}

eBay's GenAI features:
\begin{itemize}
\item Chat with eBay.ai*
\item eBay AI Description Generator*
\item eBay Shop the Look*
\item \textit{Three eBay features that were not yet public (one seller-side and two buyer-side, of which one buyer-side feature* was included in the Heuristic Evaluation)} 
\end{itemize}

Did not review, as they did not have dedicated feedback mechanisms as of March 2025:
\begin{itemize}
\item Amazon Customers Say
\item DeepAI
\item eBay Background Generator
\item Poshmark Smart List AI
\item Snapchat AI Snaps/AI Lenses/AI Memories
\item Tripadvisor AI
\item Walmart Customer Service
\item Yelp AI Assistant
\end{itemize}

\section{Template for Heuristic Evaluation}
\label{app:HE_template}


\textbf{Instructions:}\smallskip

\begin{mdframed}
Welcome! Because you're considered an expert in your field, you're invited to participate in a heuristic evaluation of the user interface designs and flows that allow users to provide feedback about GenAI systems' outputs. 

\textbf{What is Heuristic Evaluation?} A heuristic evaluation is a method for identifying design problems in a user interface. Evaluators judge the design against a set of guidelines (called heuristics) that make systems easy to use.

\textbf{Why are user feedback features important?} User feedback is the communication channel between the users, the product, and the product team. For AI products, user feedback and control are critical to improving the underlying AI model’s output and user experience. When users have the opportunity to offer feedback, they can play a direct role in personalizing their experiences and maximizing the benefit the product brings to them. When users have the right level of control over the system, they’re more likely to trust it.

\textbf{Task: Fill out all empty rows and columns. It will take about 20 minutes per tab.
}
\begin{enumerate}
    \item Assume that you're the user of the AI system and you want to provide positive or negative feedback about this AI system.
    \item \textbf{Watch video recordings} of the interactions with the user feedback feature for both Web and Mobile versions. You can re-watch, pause and rewind the videos. 
    \textit{Web}: [link]
    \textit{Mobile}:[link]
    \item Record your \textbf{Job role}, the \textbf{Date of evaluation}, \textbf{Device, and Operating System} 
    on which the user feedback was tested. 
    \item Read the description of each \textbf{Heuristic/Criterion}. 
    \item Evaluate the user feedback feature by answering all \textbf{Guiding Questions}; note that it is not an exhaustive list; feel free to share any additional thoughts. In the \textbf{Rating} column, provide an overall assessment of how well the Heuristic is implemented in the user feedback feature. 
    \item (most important) Provide comments about identified \textbf{Issues}, note \textbf{Positive aspects}, and suggest \textbf{Recommendations} for improvement. Be as specific as you can.
\end{enumerate}
\end{mdframed}

Table \ref{tab:HE_template} demonstrates the template format provided to the evaluators, and Table \ref{tab:heuristics} gives the full list of heuristics and guiding questions.


\begin{longtable}{p{4cm}p{4.2cm}|p{1.4cm}|p{0.7cm}|p{1.1cm}|p{1.4cm}}
\caption{Template for heuristic evaluation, with heuristic, guiding questions, and space to record rating and written evaluation. Only the first 2 heuristics are included in this format for illustrative purposes. The full list of heuristics is in Table~\ref{tab:heuristics}.} 
\label{tab:HE_template} \\  															

\toprule
\textbf{Heuristic/Criterion} & 
\textbf{Guiding Questions} 	& 
\textbf{Rating}  & 
\textbf{Issues} &\textbf{Positive aspects}	& \textbf{Recommen-dations} \\
\midrule
\endfirsthead  
\multicolumn{6}{c}{{\bfseries \tablename\ \thetable{} -- \textit{continued from previous page}}} \\
\toprule
\textbf{Heuristic/Criterion} & \textbf{Guiding Questions} 	& \textbf{Rating} & \textbf{Issues} &	\textbf{Positive aspects}	& \textbf{Recommen-dations} \\
\midrule
\endhead  
\midrule \multicolumn{6}{r}{{\textit{Continued on next page}}} 															  \\ \midrule 
\endfoot  
\bottomrule \multicolumn{6}{r}{{\textit{End of table}}} 															  \\ \midrule 
\endlastfoot  

\textbf{Entry Point:} The entry point should be obvious and intuitive. Users should easily understand where to start if they want to leave feedback regarding the AI-generated response.  
& 
\begin{itemize}[leftmargin=*]
    \item Is it immediately obvious that the user can leave feedback about the AI-generated response?
    \item How easy or difficult is it to discover where to start to leave feedback about the AI-generated response?
\end{itemize}
& \textit{[Very well; Somewhat well; Somewhat poorly; Very poorly]} & & & \\
\midrule  

\textbf{Navigation:} It should be easy to navigate the system to leave feedback about the AI-generated response. Information should appear in a natural and logical order. 
& 
\begin{itemize}[leftmargin=*]
    \item Does it take a lot of effort or time to leave feedback about the AI-generated response?
    \item Is information given at the points in the flow when it is most relevant to the user's decision-making?
    \item Is it obvious which form fields are optional and which ones are required?
    \item Is the interface visually appealing and easy to navigate?
\end{itemize}
& \textit{[Very well; Somewhat well; Somewhat poorly; Very poorly]} & & & \\

\bottomrule  
\end{longtable}

\begin{longtable}{p{5.2cm}p{10.3cm}}
\caption{Description of heuristics and related guiding questions used in the heuristic evaluation template.} 
\label{tab:heuristics} \\  															

\toprule
\textbf{Heuristic/Criterion} & 
\textbf{Guiding Questions} 	\\
\midrule
\endfirsthead  
\multicolumn{2}{c}{{\bfseries \tablename\ \thetable{} -- \textit{continued from previous page}}} \\
\toprule
\textbf{Heuristic/Criterion} & \textbf{Guiding Questions} \\
\midrule
\endhead  
\midrule \multicolumn{2}{r}{{\textit{Continued on next page}}} 															  \\ \midrule 
\endfoot  
\bottomrule \multicolumn{2}{r}{{\textit{End of table}}} 															  \\ \midrule 
\endlastfoot  

\textbf{Entry Point:} The entry point should be obvious and intuitive. Users should easily understand where to start if they want to leave feedback regarding the AI-generated response.  
& 
\begin{itemize}[leftmargin=*]
    \item Is it immediately obvious that the user can leave feedback about the AI-generated response?
    \item How easy or difficult is it to discover where to start to leave feedback about the AI-generated response?
\end{itemize} \\
\midrule  

\textbf{Navigation:} It should be easy to navigate the system to leave feedback about the AI-generated response. Information should appear in a natural and logical order. 
& 
\begin{itemize}[leftmargin=*]
    \item Does it take a lot of effort or time to leave feedback about the AI-generated response?
    \item Is information given at the points in the flow when it is most relevant to the user's decision-making?
    \item Is it obvious which form fields are optional and which ones are required?
    \item Is the interface visually appealing and easy to navigate?
\end{itemize} \\
\midrule  

\textbf{Clarity of the Wording/Information:} The design should speak the users' language. Use words, phrases, and concepts familiar to the user, rather than internal jargon. 
&
\begin{itemize}[leftmargin=*]
    \item Will users be familiar with the terminology used by the feedback feature? (If not, what words may be confusing?)
    \item Is the information given at each step sufficient?
    \item If there are quick-answer response options (pills, checkboxes, etc.), are they clear, relevant, and helpful to have?
\end{itemize} \\
\midrule  

\textbf{Clarity of the Process:} Users should understand the consequences of their actions when they decide to provide feedback about AI-generated responses. 
&
\begin{itemize}[leftmargin=*]
    \item Is it clear what information will be sent alongside the user's responses in the feedback form? (e.g. user ID, personal info such as email address or name, entire chat, only the AI-generated response that the user is providing feedback for, query + the response, metadata (such as time stamp, browser version, device model), etc.)
    \item Is it clear who will have access to the provided feedback? For example, would other users, buyers, or sellers be notified about the feedback? Will employees of the company that deployed the AI feature or the company that trained the model get access to the feedback? Will law enforcement be notified about any types of negative experiences reported?
\end{itemize} \\
\midrule  

\textbf{User Value and Motivation:} The benefit of providing feedback should be clear and specific. If it is not clear and specific, users may not understand why they should give feedback, they might avoid giving feedback, or if they can’t avoid it, they could give meaningless responses, or feedback that ends up being harmful to the product or community.
&
\begin{itemize}[leftmargin=*]
    \item Is it clear what will happen if the user submits the feedback? For example, will future AI-generated responses change for the specific user (i.e. they will not see similar responses anymore or recommendations will be more customized)? Will they change for all users (e.g. it will be used in re-training or fine-tuning of the model in general)? Will anyone (e.g. other users, buyers, sellers) be penalized if users report negative experiences with the AI-generated responses?
    \item Is the tone of the communication friendly, respectful, and inviting? Will the user feel encouraged to submit feedback?
    \item Does the flow include a thank you message to make users feel that their feedback is appreciated?
\end{itemize} \\
\midrule  

\textbf{Help and Additional Information:} It’s best if the system doesn’t need any additional explanation. However, it may be necessary to provide documentation to help users understand how to complete their tasks.
&
\begin{itemize}[leftmargin=*]
    \item Is help or additional information easy to find?
    \item Is help or additional information provided in context right at the moment when the user requires it? (For example, via hover-over text, in-situ explanations, ""Learn more"" links.)
\end{itemize} \\
\midrule  

\textbf{Unnecessary Information:} While interfaces should contain sufficient information, they also should not contain information that is irrelevant or rarely needed (at least within the main flow). Every extra unit of information in an interface competes with the relevant units of information and diminishes their relative visibility.
&
\begin{itemize}[leftmargin=*]
    \item Is the visual design and content focused on the essentials?
    \item Have all distracting, unnecessary elements been removed?
\end{itemize} \\
\midrule  

\textbf{Consistency:} The experience should be consistent across device types and interaction channels. Users should not have to wonder whether different words, situations, or actions mean the same thing. The feature should follow platform and industry conventions.
&
\begin{itemize}[leftmargin=*]
    \item Are words and visual elements used consistently throughout the flow?
    \item Are words and visual elements used consistently across different types of devices or different channels (websites vs. mobile apps), when feasible?
    \item Does the design follow industry conventions?
\end{itemize} \\
\midrule  

\textbf{User Control and Freedom:} Users often perform actions by mistake, or open menus just to explore what's there when looking for something. They need a clearly marked ""emergency exit"" to leave an unwanted action, without having to go through an extended process.
&
\begin{itemize}[leftmargin=*]
    \item Does the design allow users to go back a step in the process?
    \item Are ``exit'' options easily discoverable? Is it clear how to cancel the action if the user decides not to submit the feedback?
    \item Are Undo and Redo supported?
\end{itemize} \\
\midrule  

\textbf{Visibility of System Status:} The design should always keep users informed about what is going on, through appropriate signals from the system within a reasonable amount of time.
&
\begin{itemize}[leftmargin=*]
    \item Does the design clearly communicate its state? (For example, whether the feedback has been submitted or what rating was selected?)
    \item Do signals from the system appear immediately after user actions? Are signals persistent (i.e. do they stay on the screen or disappear)?
\end{itemize} \\
\midrule  

\textbf{Responsible AI:} The feedback form should encourage feedback not only about the relevance and factual accuracy of the response, but also other aspects of responsible and ethical AI.  
&
\begin{itemize}[leftmargin=*]
    \item Does the interface explicitly offer users a means to provide feedback about ethical aspects of the AI-generated response (e.g. bias, offensive content, or other harms)?
\end{itemize} \\
\midrule  

\textbf{Overall Assessment:} Users should feel overall satisfaction after their experience of providing feedback.
&
\begin{itemize}[leftmargin=*]
    \item Overall, how satisfied do you think an average user would be with this experience?
    \item Is there anything in this experience that really stands out as an example of good usability?
    \item Is there anything in this experience that really stands out as an example of poor usability?
\end{itemize} \\
\bottomrule  
\end{longtable}	

\clearpage

\section{Model User Feedback Template and Guidelines}
\label{appx:guidelines}



\begin{figure}[htb]
\begin{center}
\includegraphics[width=\columnwidth]{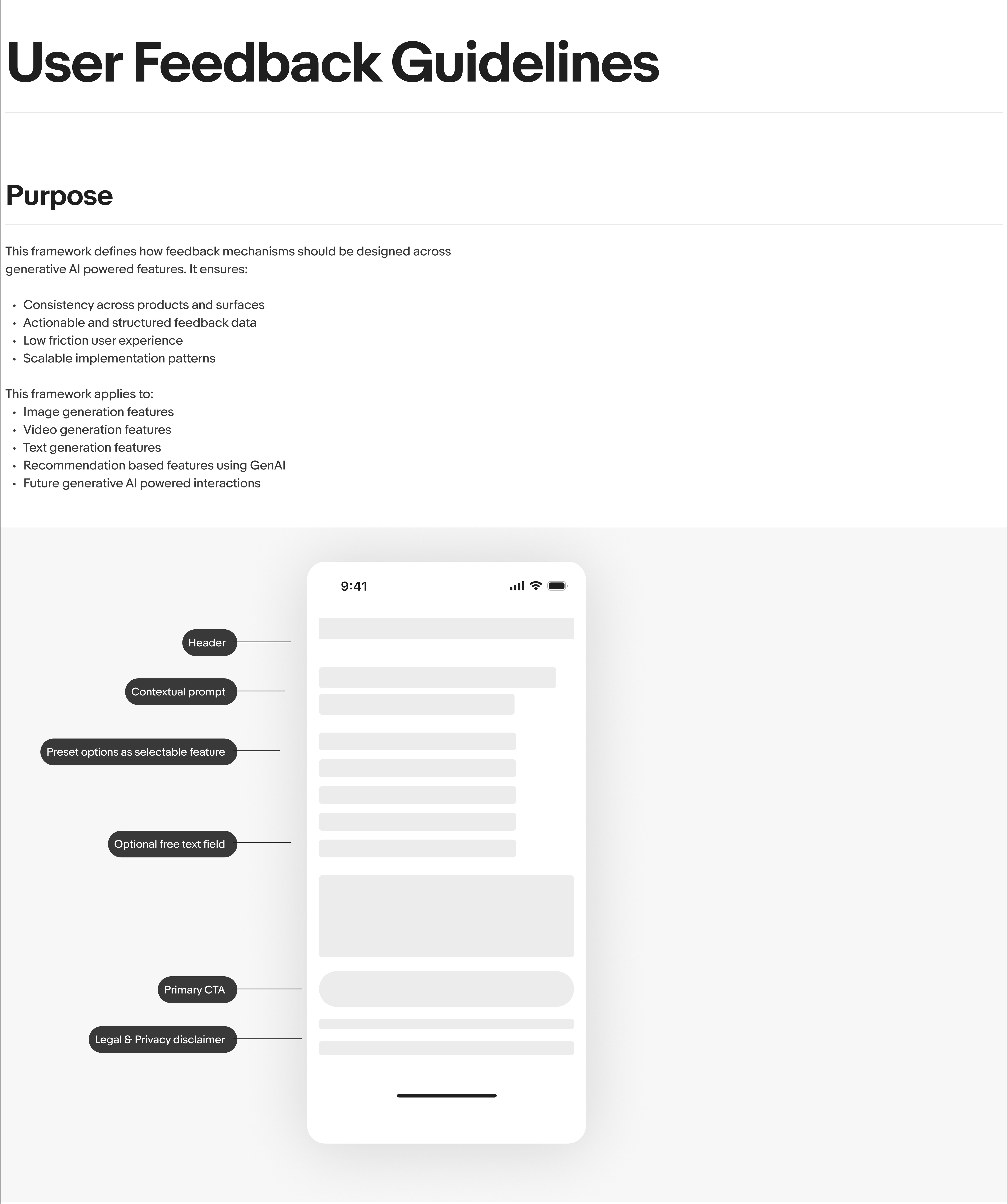} 
\end{center}
\caption{GenAI Feedback Guidelines (Part 1): Purpose of guidelines and skeleton for feedback form. }
\Description[See Supplementary Materials for text-only version of guidelines]{An easy-to-navigate text-only version of the guidelines and template is in the file `Text-Only Version of Guidelines and Template' in the Supplementary Materials for this publication.}
\label{fig:guideline_1}
\end{figure}

\begin{figure}[htb]
\begin{center}
\includegraphics[width=\columnwidth]{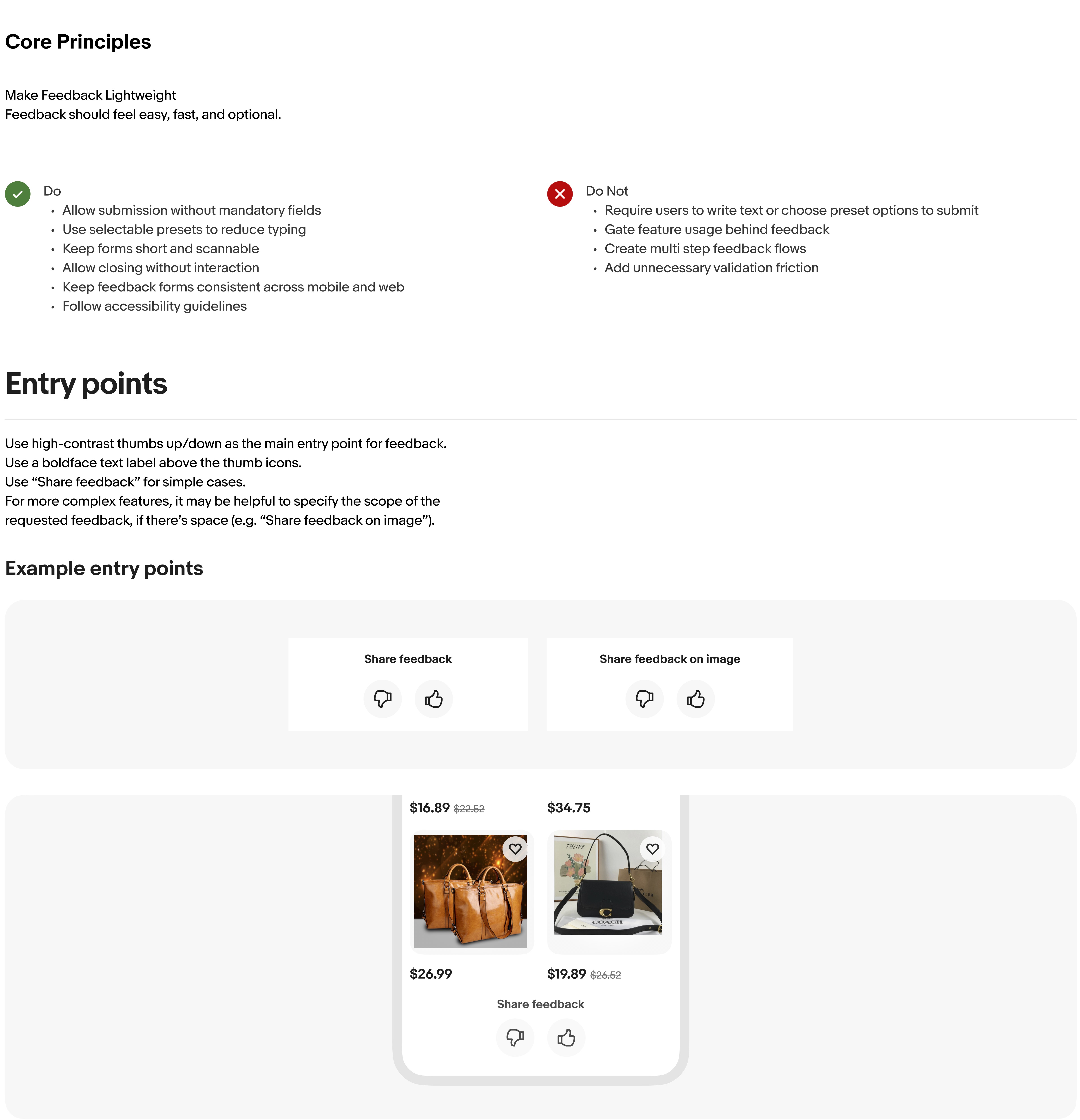} 
\end{center}
\caption{GenAI Feedback Guidelines (Part 2): Core principles and introduction to entry points.}
\Description[See Supplementary Materials for text-only version of guidelines]{An easy-to-navigate text-only version of the guidelines and template is in the file `Text-Only Version of Guidelines and Template' in the Supplementary Materials for this publication.}
\label{fig:guideline_2}
\end{figure}

\begin{figure}[htb]
\begin{center}
\includegraphics[width=\columnwidth]{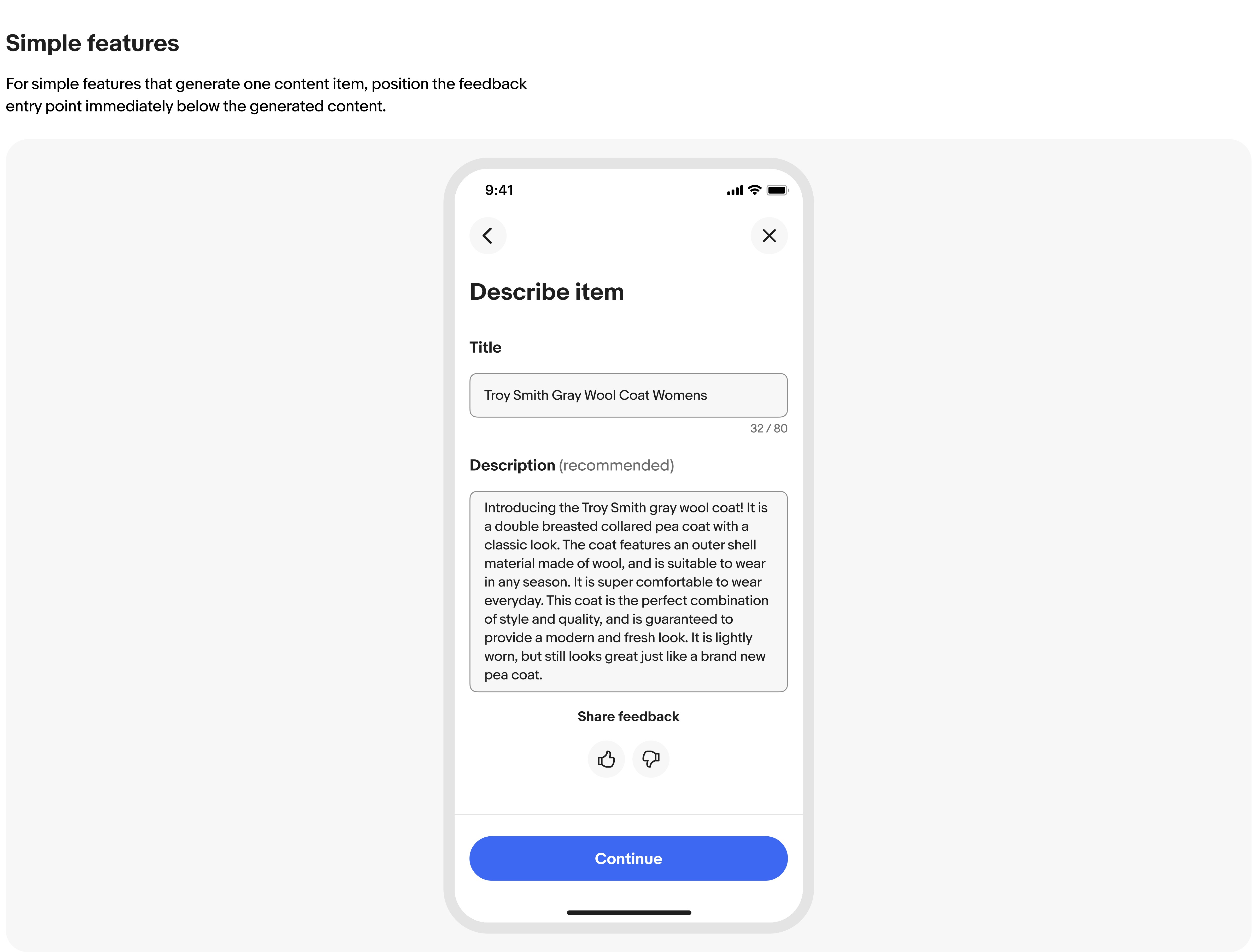} 
\end{center}
\caption{GenAI Feedback Guidelines (Part 3): Entry points for simple features.}
\Description[See Supplementary Materials for text-only version of guidelines]{An easy-to-navigate text-only version of the guidelines and template is in the file `Text-Only Version of Guidelines and Template' in the Supplementary Materials for this publication.}
\label{fig:guideline_3}
\end{figure}

\begin{figure}[htb]
\begin{center}
\includegraphics[width=\columnwidth]{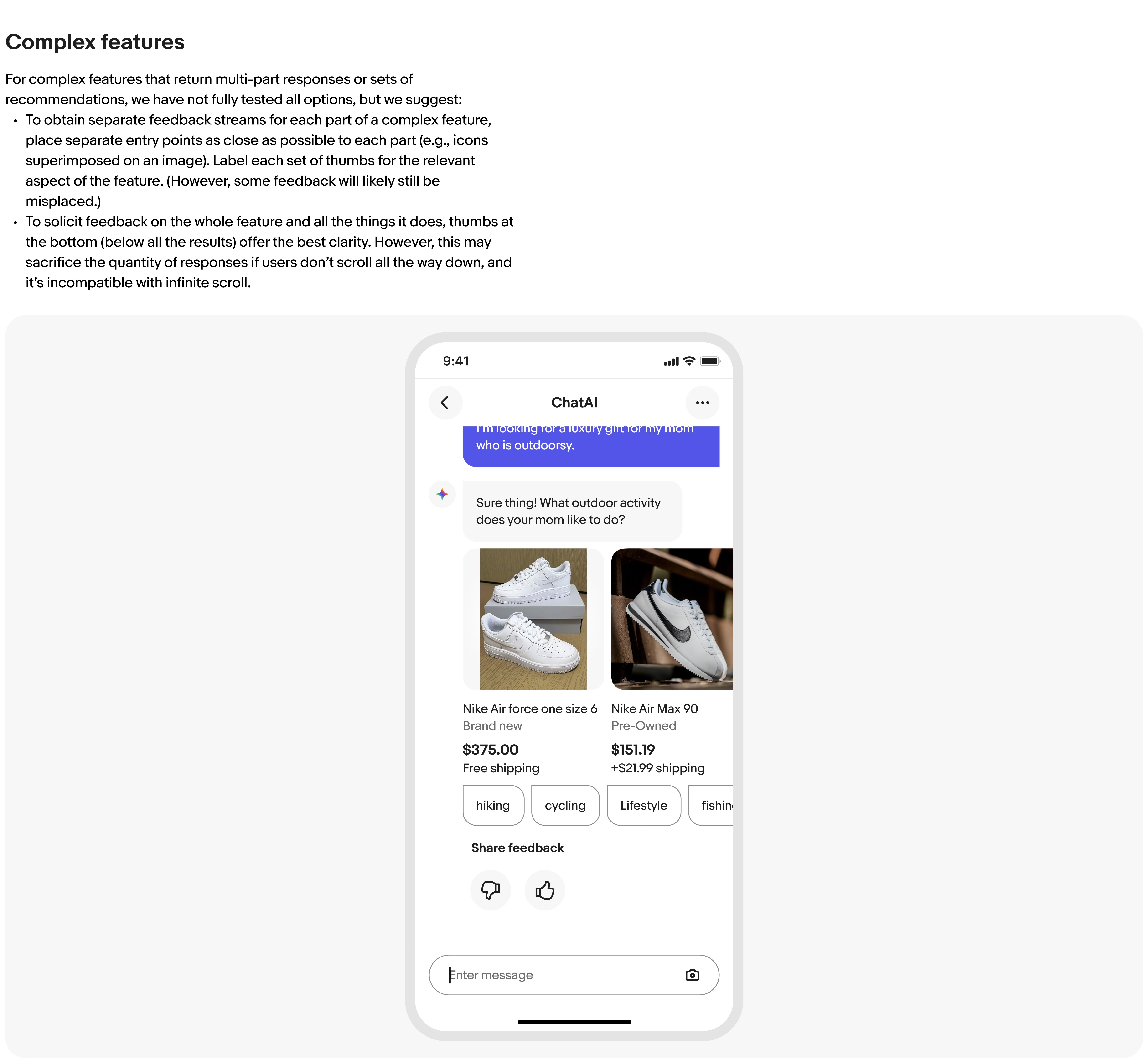} 
\end{center}
\caption{GenAI Feedback Guidelines (Part 4): Entry points for complex features.}
\Description[See Supplementary Materials for text-only version of guidelines]{An easy-to-navigate text-only version of the guidelines and template is in the file `Text-Only Version of Guidelines and Template' in the Supplementary Materials for this publication.}
\label{fig:guideline_4}
\end{figure}

\begin{figure}[htb]
\begin{center}
\includegraphics[width=\columnwidth]{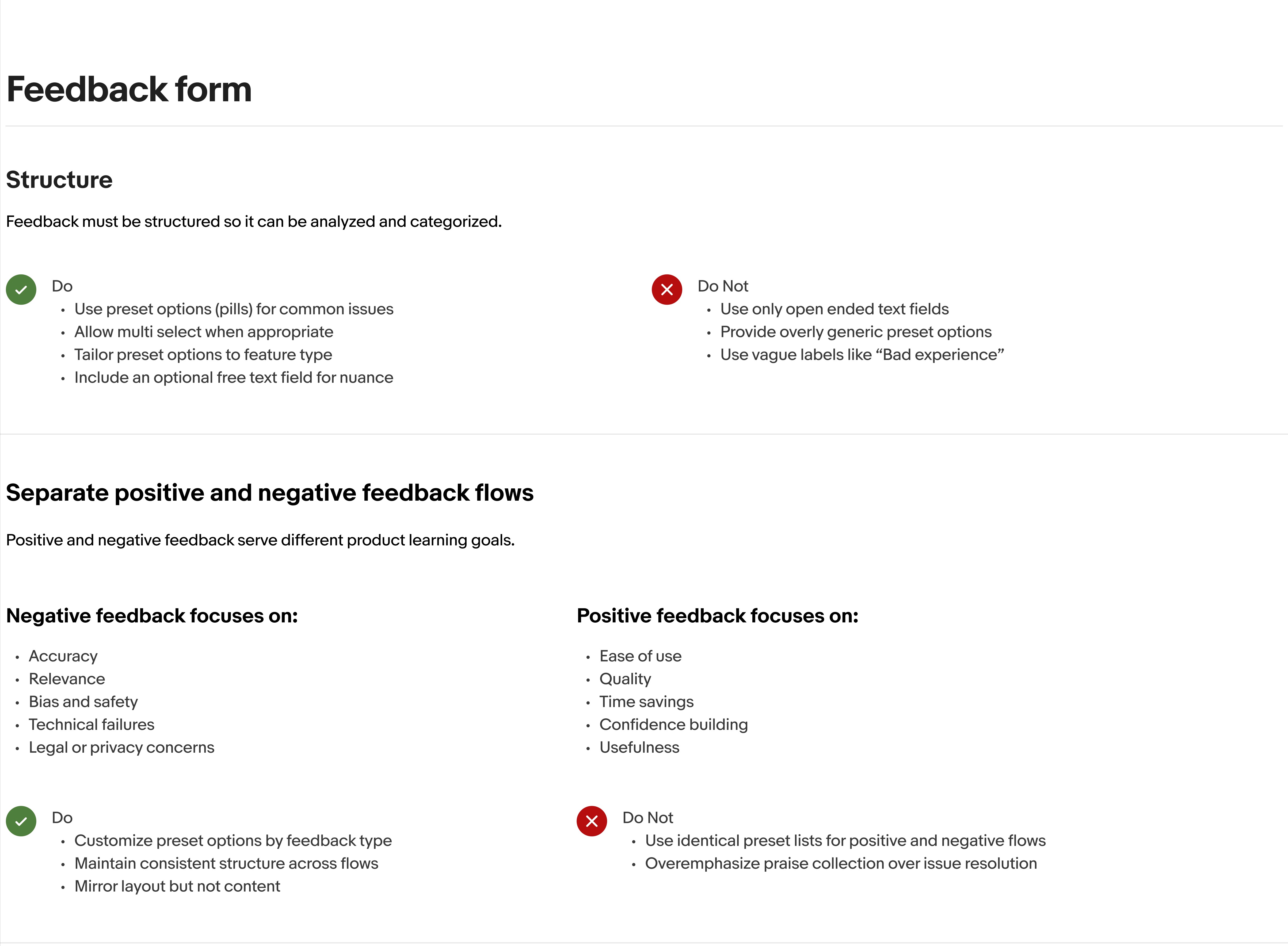} 
\end{center}
\caption{GenAI Feedback Guidelines (Part 5): Structure pointers for feedback form and for negative and positive feedback flows.}
\Description[See Supplementary Materials for text-only version of guidelines]{An easy-to-navigate text-only version of the guidelines and template is in the file `Text-Only Version of Guidelines and Template' in the Supplementary Materials for this publication.}
\label{fig:guideline_5}
\end{figure}

\begin{figure}[htb]
\begin{center}
\includegraphics[width=\columnwidth]{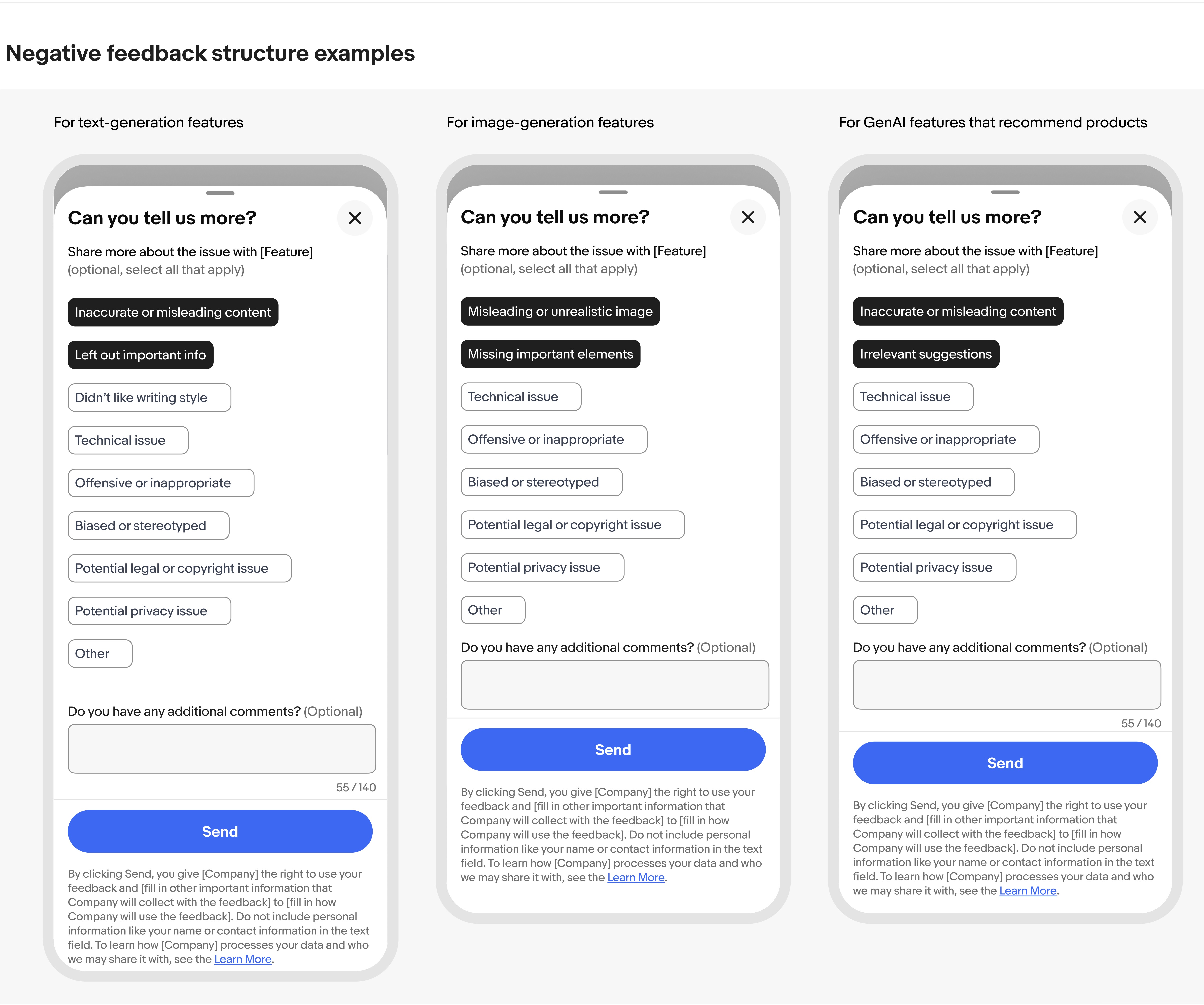} 
\end{center}
\caption{GenAI Feedback Guidelines (Part 6): Negative feedback forms for different types of features.}
\Description[See Supplementary Materials for text-only version of guidelines]{An easy-to-navigate text-only version of the guidelines and template is in the file `Text-Only Version of Guidelines and Template' in the Supplementary Materials for this publication.}
\label{fig:guideline_6}
\end{figure}

\begin{figure}[htb]
\begin{center}
\includegraphics[width=\columnwidth]{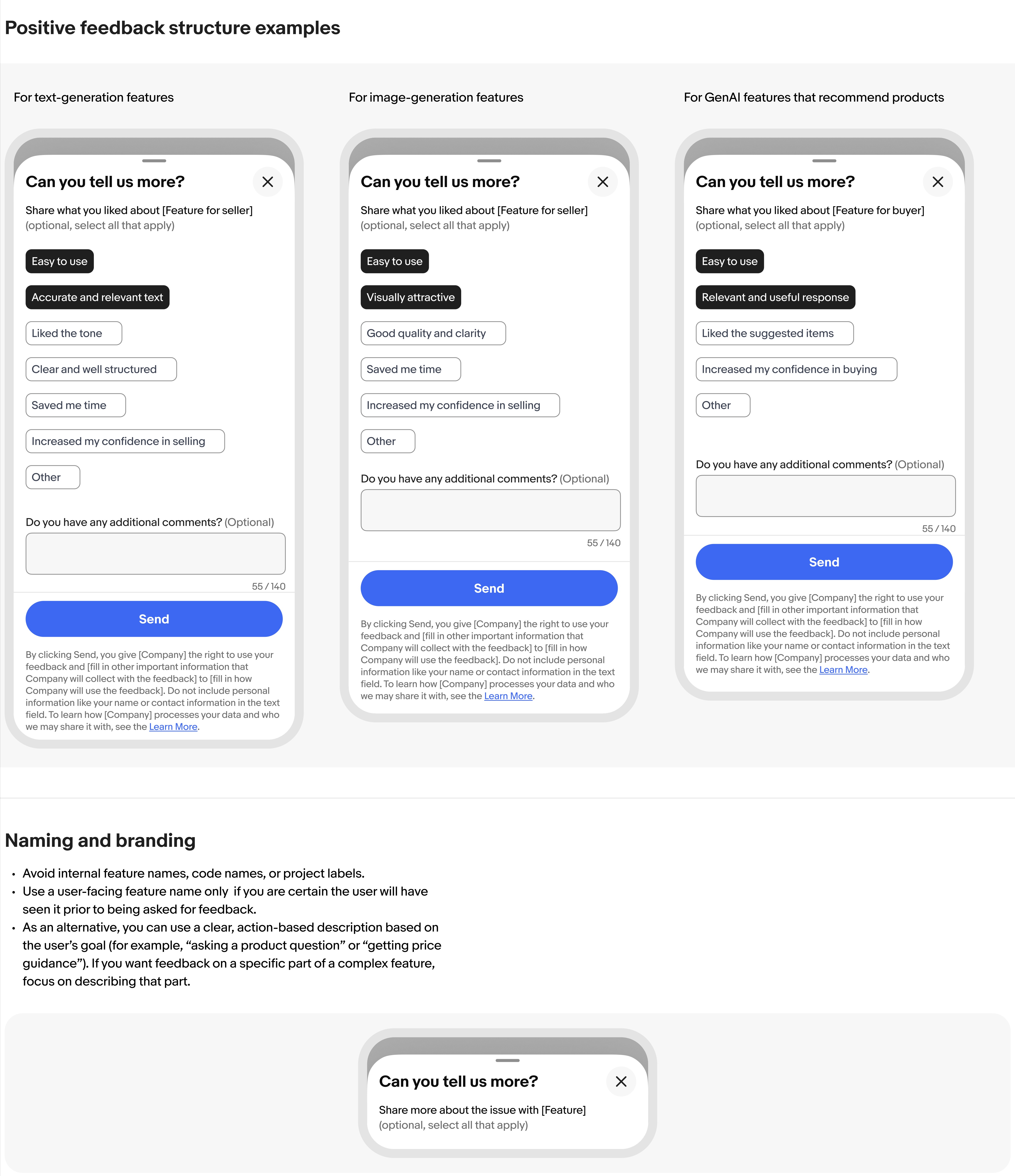} 
\end{center}
\caption{GenAI Feedback Guidelines (Part 7): Positive feedback forms for different types of features.}
\Description[See Supplementary Materials for text-only version of guidelines]{An easy-to-navigate text-only version of the guidelines and template is in the file `Text-Only Version of Guidelines and Template' in the Supplementary Materials for this publication.}
\label{fig:guideline_7}
\end{figure}

\begin{figure}[htb]
\begin{center}
\includegraphics[width=\columnwidth]{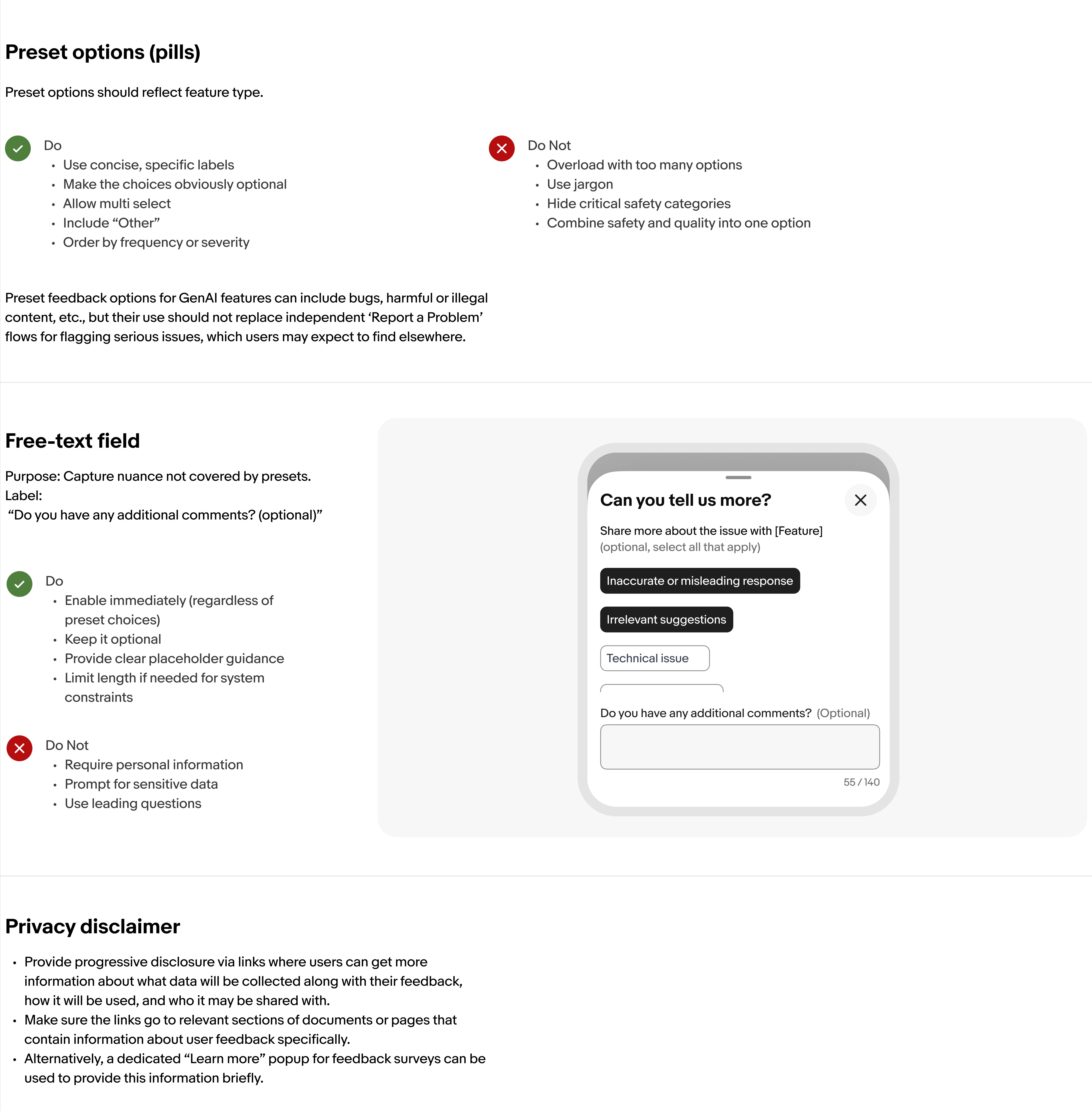} 
\end{center}
\caption{GenAI Feedback Guidelines (Part 8): Detailed pointers for preset options, free-text responses, and privacy disclaimers.}
\Description[See Supplementary Materials for text-only version of guidelines]{An easy-to-navigate text-only version of the guidelines and template is in the file `Text-Only Version of Guidelines and Template' in the Supplementary Materials for this publication.}
\label{fig:guideline_8}
\end{figure}

\begin{figure}[htb]
\begin{center}
\includegraphics[width=\columnwidth]{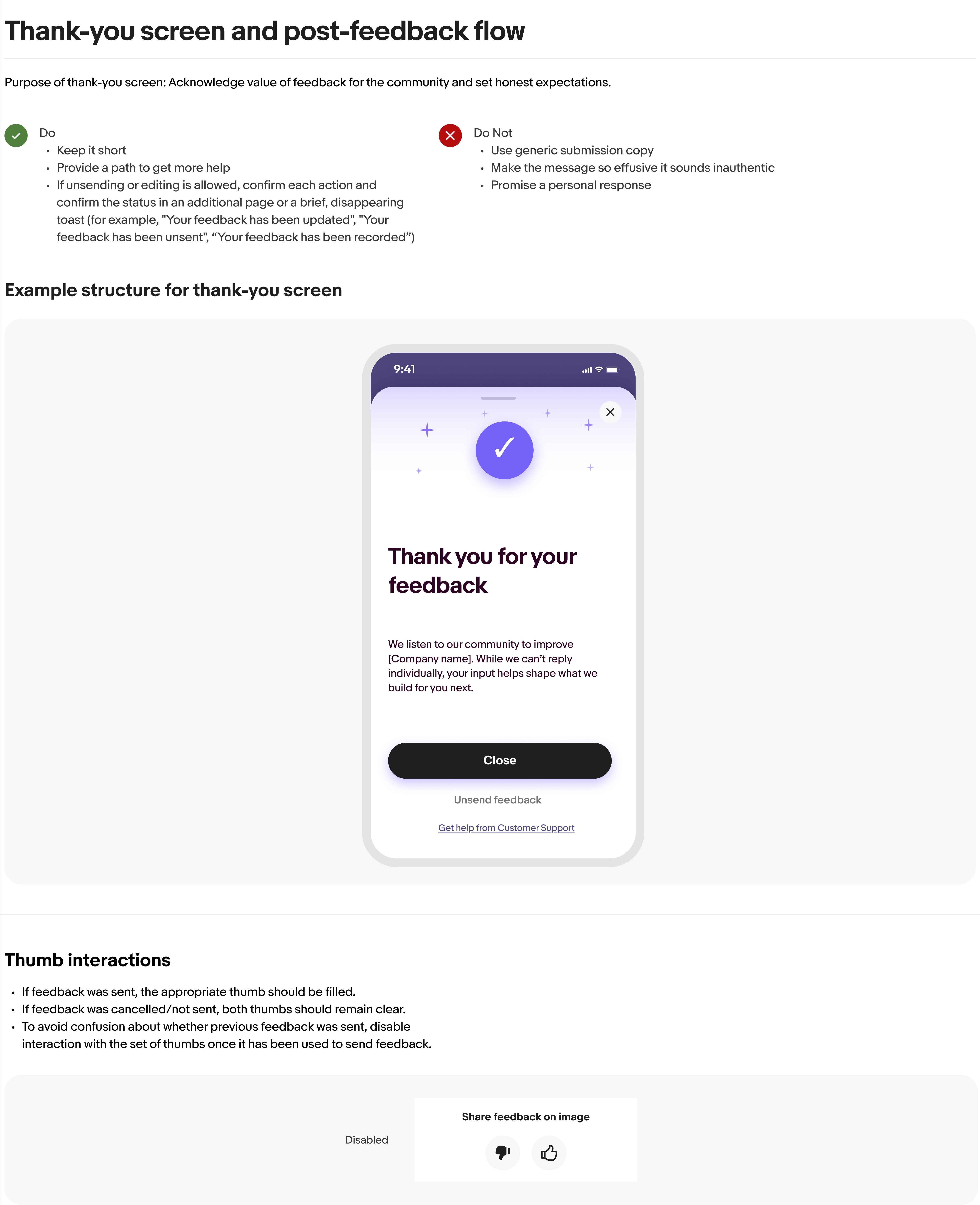} 
\end{center}
\caption{GenAI Feedback Guidelines (Part 9): Thank-you screens and post-feedback signals.}
\Description[See Supplementary Materials for text-only version of guidelines]{An easy-to-navigate text-only version of the guidelines and template is in the file `Text-Only Version of Guidelines and Template' in the Supplementary Materials for this publication.}
\label{fig:guideline_9}
\end{figure}

Figures \ref{fig:guideline_1}--\ref{fig:guideline_9} show the final set of guidelines for user feedback on GenAI tools and features, incorporating findings and recommendations from all stages of the project.

These guidelines are available in alternative formats in the Supplementary Materials:
\begin{itemize}
\item A text-only version of the guidelines and text for the template, organized suitably for screenreaders, in the file \textit{Text-Only Version of Guidelines and Template}.
\item An single image PDF of the full guidelines and template in color, in the file \textit{Full-Image Version of Guidelines and Template}.
\end{itemize}

\end{document}